\documentclass[journal]{IEEEtran}

\usepackage{soul}

\usepackage{cite}

\usepackage{booktabs}

\ifCLASSINFOpdf
  \usepackage[pdftex]{graphicx}
\else
\fi
\usepackage{amsmath}
\usepackage{amsfonts}
\usepackage{cases}

\usepackage{algorithm}
\usepackage{algorithmicx}
\usepackage{algpseudocode}

\ifCLASSOPTIONcompsoc
 \usepackage[caption=false,font=normalsize,labelfont=sf,textfont=sf]{subfig}
\else
 \usepackage[caption=false,font=footnotesize]{subfig}
\fi

\usepackage{xcolor}

\usepackage[nolist,nohyperlinks]{acronym}
\begin{acronym}
    \acro{RTF}{real time factor}
    \acro{CC}{constant curvature}
    \acro{RL}{reinforcement learning}
    \acro{DoF}{degree of freedom} 
    \acroplural{DoF}[DoFs]{degrees of freedom} 
    \acro{IL}{imitation learning}
    \acro{LfD}{learning from demonstration}
    \acro{UJ}{universal joint}
    \acroplural{UJ}[UJs]{universal joints}
\end{acronym}

\usepackage[hidelinks]{hyperref}
\usepackage{refcount}

\usepackage{censor}
\StopCensoring

\ifcensor
  \newcommand{\balooMujocoSimUrl}{https://example.com/censored-link1}
  \newcommand{\balooGymUrl}{https://example.com/censored-link2}
  \newcommand{\videoUrl}{https://youtu.be/MvORMfDxfq0}

\else
  \newcommand{\balooMujocoSimUrl}{https://github.com/byu-rad-lab/baloo-mujoco-sim}
  \newcommand{\balooGymUrl}{https://github.com/byu-rad-lab/baloo-gym}
  \newcommand{\videoUrl}{https://www.youtube.com/watch?v=fv-1zDK8Wo0}
\fi

\begin{document}

\title{Zero-shot Whole-Body Manipulation with a Large-Scale Soft Robotic Torso via Guided Reinforcement Learning}
%
%
%

\author{%
\ifcensor
    \blackout{
    Curtis C. Johnson,
    Carlo Alessi,
    Egidio Falotico,
    and Marc D. Killpack
    \thanks{CCJ and MDK are with the Department of Mechanical Engineering, Brigham Young University, Provo, Utah, USA}%
    \thanks{CA is with the Humanoid Sensing and Perception, Italian Institute of Technology, Via Morego 30, 16163 Genoa, Italy}%
    \thanks{EF is with The BioRobotics Institute, Scuola Superiore Sant’Anna, 56025 Pisa, Italy}
    }%
\else
    Curtis C. Johnson,
    Carlo Alessi,
    Egidio Falotico,
    and Marc D. Killpack
    \thanks{CCJ and MDK are with the Department of Mechanical Engineering, Brigham Young University, Provo, Utah, USA}%
    \thanks{CA is with the Humanoid Sensing and Perception, Italian Institute of Technology, Via Morego 30, 16163 Genoa, Italy}%
    \thanks{EF is with The BioRobotics Institute, Scuola Superiore Sant’Anna, 56025 Pisa, Italy}%
\fi
}

%
%

\markboth{Journal of \LaTeX\ Class Files,~Vol.~14, No.~8, August~2015}%
{Shell \MakeLowercase{\textit{et al.}}: Bare Demo of IEEEtran.cls for IEEE Journals}
%



\maketitle

\begin{abstract}
Whole-body manipulation is a powerful yet underexplored approach that enables robots to interact with large, heavy, or awkward objects using more than just their end-effectors. Soft robots, with their inherent passive compliance, are particularly well-suited for such contact-rich manipulation tasks, but their uncertainties in kinematics and dynamics pose significant challenges for simulation and control. In this work, we address this challenge with a simulation that can run up to 350x real time on a single thread in MuJoCo and provide a detailed analysis of the critical tradeoffs between speed and accuracy for this simulation. Using this framework, we demonstrate a successful zero-shot sim-to-real transfer of a learned whole-body manipulation policy, achieving an 88\% success rate on the \blackout{Baloo} hardware platform. We show that guiding \ac{RL} with a simple motion primitive is critical to this success where standard reward shaping methods struggled to produce a stable and successful policy for whole-body manipulation. Furthermore, our analysis reveals that the learned policy does not simply mimic the motion primitive. It exhibits beneficial reactive behavior, such as re-grasping and perturbation recovery. We analyze and contrast this learned policy against an open-loop baseline to show that the policy can also exhibit aggressive over-corrections under perturbation. To our knowledge, this is the first demonstration of forceful, six-\ac{DoF} whole-body manipulation using two continuum soft arms on a large-scale platform (10~kg payloads), with zero-shot policy transfer.
\end{abstract}


%
\IEEEpeerreviewmaketitle

\section{Introduction}
\IEEEPARstart{S}{oft} robots have garnered significant attention in recent years due to their potential to interact with unstructured environments in a natural and safe manner without the need for high bandwidth control and sensing \cite{rus2015design}. Because of their passive compliance, soft robots are particularly well-suited for contact-rich manipulation of objects that are traditionally difficult to handle due to uncertainty, size, geometry, or weight. While research efforts--with both rigid and soft robots--have historically focused on interacting with the environment with an end effector, we emphasize that a more flexible view of manipulation can be highly beneficial. A type of manipulation where a robot interacts with the environment using any part of its structure is called whole-arm or whole-body manipulation. When lifting large, heavy objects, humans naturally engage their whole arms and torso to achieve a stable grasp--something beyond the capabilities of using just our hands. If a robot is able to use more than its end-effectors to interact with the world, the capabilities of the robot can expand dramatically, especially with regard to manipulating large, bulky, heavy, or awkward objects. Some practical applications that could substantially benefit from these expanded capabilities include assistive robotics, industrial material handling, disaster cleanup, or rescue operations. By leveraging the benefits of soft robotic interactions for whole-body manipulation, we are able to interact with objects that would otherwise be impossible to manipulate.

\begin{figure}[t]
    \centering
    \includegraphics[width=\columnwidth]{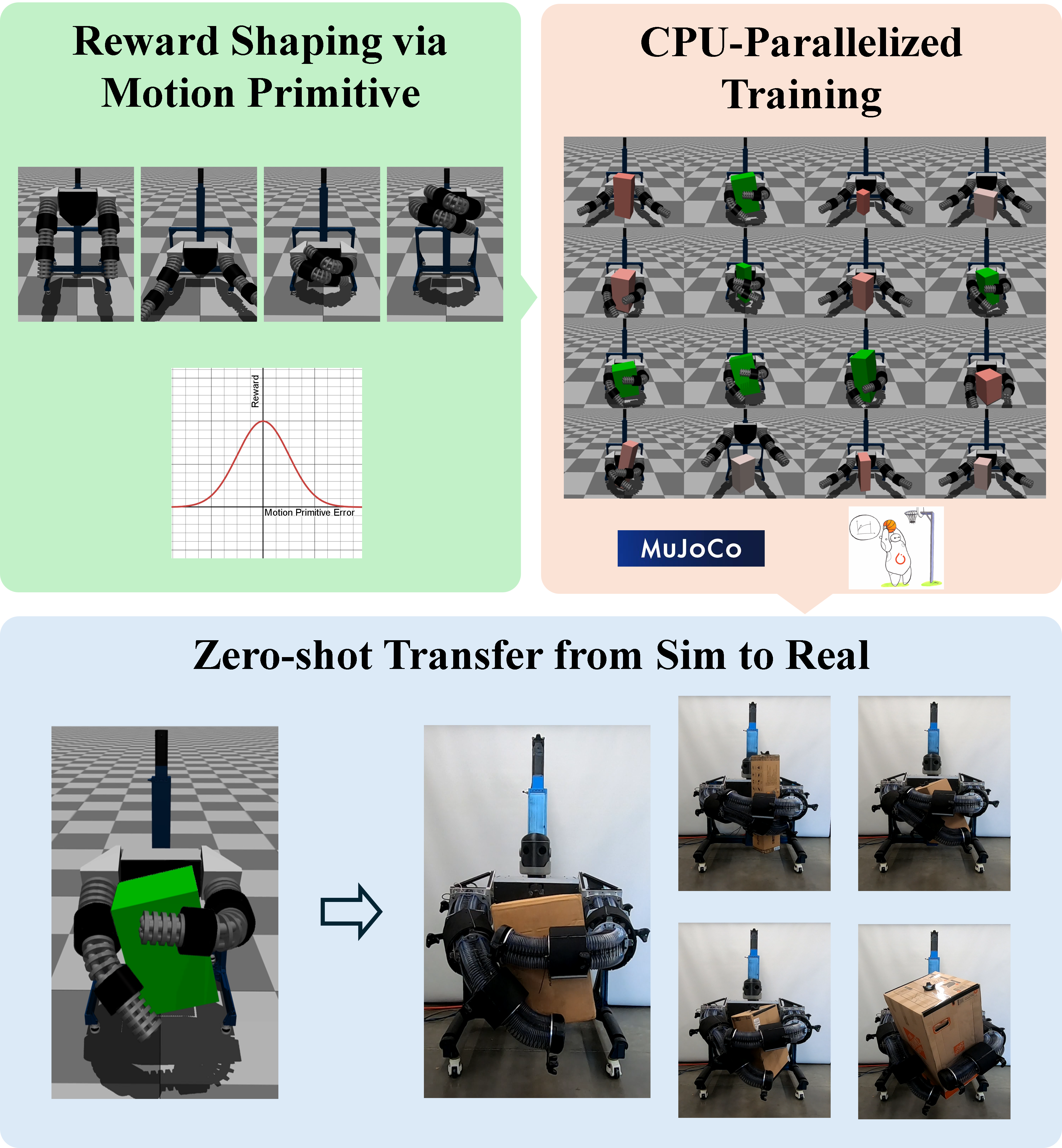}
    \caption{Overview of contributions presented in this paper. We used a shaped reward to guide the learning process with a grasping motion primitive. To learn a policy, we implemented a parallelized learning environment using MuJoCo and StableBaselines3. After training on a set of boxes of different sizes and mass we achieved a successful zero-shot transfer of the learned policy to soft robot hardware for five different test boxes.}
    \label{fig:baloo-rl}
\end{figure}

However, the use of soft robots does not come without disadvantages. Soft robots often exhibit highly nonlinear dynamics as well as significant hysteresis and uncertainty that make precise manipulation tasks difficult. Control strategies for soft robot manipulation most often employ combinations of open-loop trajectories, model-based control algorithms, and learning \cite{wang2022control}. Open-loop trajectories are usually hand-designed for a specific task or generated from a planning algorithm. While model-based control has been used successfully with both kinematic and dynamic models \cite{della2023model}, it is well known that the success of these methods relies on accurate models of the robot and its environment. Unfortunately, finding accurate models is difficult to accomplish with soft robots.

This challenge is commonly called the sim-to-real gap. Small modeling inaccuracies can cause learned policies to fail when deployed on real hardware. For soft robots, the gap can be especially pronounced due to simplified models, nonlinearities, and history-dependent dynamics.

A substantial amount of work has explored learning-based control strategies for soft robots \cite{falotico2024learning}. Using data to learn a policy, a model, or to improve a baseline controller can reduce the effect of unmodeled disturbances and uncertainties. The obvious tradeoff is the need for data. Learning approaches, especially those involving neural networks, are often `data-hungry' \cite{ibarz2021train}. While significant progress has been made towards improving the data efficiency of these algorithms, it is common to need millions of data points to achieve satisfactory modeling performance. Real-world data is sometimes easier to obtain on soft robots relative to rigid ones because of passive compliance and soft components, but the data collection process can still cause unwanted wear. If real-world data collection is impossible or undesirable, simulation can help meet the demand for data generation. Unfortunately, developing a model of a soft robot that is both accurate enough to bridge the sim-to-real gap and fast enough to be tractable for large-scale data collection is challenging. 

We approach this challenge with a soft robot simulation framework built on top of MuJoCo \cite{todorov2012mujoco} (a high-performance rigid-body simulator) and demonstrate that our implementation is both accurate enough for zero-shot sim-to-real transfer of a learned policy and fast enough for data-intensive learning tasks. To highlight the benefits of soft robots in contact-rich manipulation, we focus on learning the task of whole-body manipulation of unknown objects. In summary, we present three main contributions summarized in Figure~\ref{fig:baloo-rl} and listed here:

\begin{enumerate}
    \item \textbf{A simulation for contact-rich manipulation with soft robots in MuJoCo:} We release our open-source simulation framework\footnote{\url{\balooMujocoSimUrl}} which runs up to 350x real time on a single thread and provide a detailed analysis quantifying critical tradeoffs between speed and accuracy. This simulation enables tractable \ac{RL} and successful zero-shot transfer of learned policies.
    \item \textbf{A successful zero-shot sim-to-real transfer of a learned whole-body grasping policy with soft robots:} Using our open-source RL pipeline\footnote{\url{\balooGymUrl}}, we achieve an 88\% hardware success rate without fine-tuning. To the best of our knowledge, this is the first demonstration of forceful whole-arm coordination between two soft robot arms at scale. Guiding \ac{RL} with a motion primitive is critical to solving this complex whole-body manipulation task, where shaped rewards alone fail.
    \item \textbf{An in-depth analysis of the effects of guided \ac{RL}:} We show that guided learning transforms an open-loop primitive into a reactive, closed-loop policy with fundamental differences from the original primitive. While it enables advanced behaviors such as re-grasping on hardware, it can also exhibit over-corrections under perturbations.
\end{enumerate}

We acknowledge recent compelling work in \cite{Barreiros_Onol_Zhang_Creasey_Goncalves_Beaulieu_Bhat_Tsui_Alspach_2025}, which similarly presents an example-guided \ac{RL} pipeline and offers further encouraging results for leveraging expert demonstrations to guide learning for whole-body robotic manipulation. Their approach utilizes both teleoperated and planned trajectories as motion primitives while using both passive and active compliance control on a bimanual rigid robot with soft inflatable shells. Furthermore, their reward structure, comprising eight distinct terms for the task reward and an imitation reward based on a discriminator from the Adversarial Motion Priors algorithm, avoids the need for designing explicit imitation objectives. This work, like ours, demonstrates the immense potential of integrating examples into RL.

Our work complements \cite{Barreiros_Onol_Zhang_Creasey_Goncalves_Beaulieu_Bhat_Tsui_Alspach_2025} by exploring guided \ac{RL} with a purely soft robot, focusing solely on the passive compliance of the robot itself. By taking compliance a step further, we avoid issues like torque-outs, which the authors of \cite{Barreiros_Onol_Zhang_Creasey_Goncalves_Beaulieu_Bhat_Tsui_Alspach_2025} noted as a challenge. Our approach to reward design also differs. We also utilize a task reward and imitation reward but out task reward is only a single term and our imitation reward directly rewards the agent in action space, which required explicit design. We also extend the applicability of example-guided \ac{RL} to a broader range of object sizes (0.2-1.1 m) and weights (1.0-9.8 kg) compared to their objects (0.15-.57 m, 0.55-1.8 kg). Because we are using a soft robot, our policy operates more end-to-end by choosing 25 low-level pressure commands as opposed to 14 higher level joint commands. Our task also requires learning how to approach the object laying on the floor instead of the object being placed on a table. Lastly, their analysis focused on the effects of active vs passive compliance and the quality of the example motions. Our analysis instead delves deeply into the fundamental differences between example motion primitives and learned behaviors. These complementary but distinct results collectively strengthen the argument for the effectiveness of compliance and guiding examples in learning whole-body manipulation policies. 

The remainder of this work is organized as follows. Section~\ref{sec:related-works} discusses work related to this paper and how our contributions fit into the current state-of-the-art. Section~\ref{sec:hardware-description} summarizes the hardware used for experiments. Section~\ref{sec:simulation-description} presents our MuJoCo simulation framework, while Section~\ref{sec:env-description} details the whole-body manipulation \ac{RL} task. Section~\ref{sec:results} discusses experimental results from both simulation and hardware. Section~\ref{sec:conclusion} concludes with a summary, followed by a discussion of current limitations and directions for future work.

\section{Related Work}
\label{sec:related-works}

In this section, we provide an overview of the state of the art related to the contributions of this paper. These include simulation methods for soft robots, using learning algorithms to control them, and coordinated, forceful control between multiple soft robots.

\subsubsection{Modeling and Simulation of Soft Robots}
Soft robots can be modeled with various mathematical techniques \cite{armanini2023soft}. Each method has tradeoffs between accuracy and computational speed. The well-studied \ac{CC} methods approximate the motion of soft robots with parametrized geometrical curves, although they are less suitable in the case of significant external contacts \cite{webster2010design}. Finite element methods, on the other hand, can offer more accurate simulations but are often computationally expensive \cite{faure2012sofa, Coevoet_Morales-Bieze_Largilliere_Zhang_Thieffry_Sanz-Lopez_Carrez_Marchal_Goury_Dequidt, dubied2022sim}. 
Cosserat rod models are a good compromise to parametrize the deformations of soft manipulators, even when subject to manufacturing irregularities \cite{alessi2023ablation}. However, the assumptions necessary for the formulation of Cosserat rod models only apply to slender robots.

In contrast, we employ a lumped-parameter approach that can be used with continuum robots of any shape and size. Different versions of lumped-parameter models have previously been explored.\cite{Katzschmann_Santina_Toshimitsu_Bicchi_Rus_2019} presents a theoretically exact match to piecewise \ac{CC} soft robots. However, the method requires computationally expensive nonlinear mapping between the parameterized configuration space and curvature space. The method also does not generalize to non-slender continuum robots. SoMo \cite{graule2021somo} uses PyBullet to simulate a serial chain of rigid bodies interconnected with spring-loaded joints. The simulation includes a mapping between parameterized configuration space and curvature space. We adapt this concept to build the simulation in MuJoCo since it is well-supported, actively-developed, and much faster than PyBullet. 

\cite{graule2021somo} reports simulation speeds of up to 0.5x real time in SoMo. The SOFA framework \cite{Coevoet_Morales-Bieze_Largilliere_Zhang_Thieffry_Sanz-Lopez_Carrez_Marchal_Goury_Dequidt} has also been used successfully to simulate soft continuum robot manipulation but \cite{Schegg_Ménager_Khairallah_Marchal_Dequidt_Preux_Duriez_2023} reports speeds from .007 to 18.6x real time across all tasks. Our simulation framework can achieve speeds up to 350x real time. \cite{Morimoto_Ikeda_Niiyama_Kuniyoshi_2022} and \cite{Morimoto_Nishikawa_Niiyama_Kuniyoshi_2021} also use MuJoCo to simulate a continuum robot arm, but only provide minimal implementation details. While approximating continuum robots with serial chains of rigid links is an established technique, our primary contribution is a high-performance and open-source implementation of the technique and a detailed analysis of the speed-accuracy tradeoffs. The resulting simulation environment, \blackout{baloo}-mujoco-sim, is released open-source to facilitate reproducibility and future research.

\subsubsection{Learning-based Control of Soft Robots}
Machine learning approaches to control soft robots are enabling remarkable capabilities \cite{falotico2024learning}. However, most literature concerning physical interaction tasks with soft robots are limited to simulation. For instance, \cite{naughton2021elastica} trained RL policies for adaptive maneuvering through unstructured obstacles with a synthetic arm. \cite{Jitosho_Lum_Okamura_Liu_2023} used RL to train a soft robot to dynamically reach a target on a shelf where a simple PID controller fails. Also, \cite{agabiti2023whole} proposed a whole-arm grasping strategy with an elephant trunk-inspired soft arm. Recently, \cite{alessi2024pushing} achieved the sim-to-real transfer of a pose/force control policy for a dexterous soft manipulator in non-prehensile pushing tasks using deep RL and domain randomization. Also, a data-efficient Bayesian optimization-based approach was proposed for learning control policies for dynamic tasks on a large-scale soft arm \cite{zwane2024learning}. Soft robotic manipulation in the real world remains, however, largely unexplored. 

Prior work, such as SofaGym \cite{Schegg_Ménager_Khairallah_Marchal_Dequidt_Preux_Duriez_2023} and SoMoGym \cite{Graule_McCarthy_Teeple_Werfel_Wood_2022}, has demonstrated learned tasks like pen spinning, in-hand manipulation, and object reorientation using small soft robots or grippers. However, all of the tasks learned in \cite{Schegg_Ménager_Khairallah_Marchal_Dequidt_Preux_Duriez_2023} are in simulation only (as the focus of the paper is on the software library). \cite{Graule_McCarthy_Teeple_Werfel_Wood_2022} demonstrated sim-to-real transfer for reorienting a small cube resting on a surface. In contrast, we advance soft robot control by investigating a more challenging task in simulation and on hardware through deep \ac{RL}: whole-body manipulation with a large-scale soft robot which has a high-dimensional action space. Unlike prior work, our task requires lifting and supporting large, heavy objects in all six \acp{DoF} instead of repositioning objects resting on a surface.

While \ac{IL} \cite{zare2024survey, ravichandar2020recent} have been successful in learning complex robotic behavior, applying these techniques to high-dimensional soft robots presents significant challenges \cite{sarker2025review}. Creating an intuitive teleoperation interface to gather expert demonstrations that are required by \ac{IL} is an open research problem. This is especially the case for non-anthropomorphic, deformable, continuum robots where mapping human behaviors to robot actions is non-trivial \cite{nazeer2023soft}. Our work explores an alternative that still leverages human intuition without requiring full demonstrations. 

\subsubsection{Coordinated Control of Multiple Soft Robots}

While soft robots have demonstrated forceful interactions with their environments (e.g., planar interaction \cite{DellaSantina_Katzschmann_Biechi_Rus_2018} and pushing \cite{alessi2024pushing}), the interaction is typically limited to single-arm manipulators and assumes contact only at the end effector. A few dual soft robot arm systems have been explored for underwater applications \cite{Ma_Monk_Cheneler_2022} and small-scale assembly tasks \cite{Wang_Xu_2019}. \cite{Oh_Rodrigue} demonstrates dual-arm lifting of several boxes weighing up to 239 grams and \cite{Russo_Sriratanasak_Ba_Dong_Mohammad_Axinte_2022} demonstrates that two continuum arms can cooperatively lift payloads of up to 272 grams. \cite{Kraus_Jensen_Killpack_2020} uses cooperative control between a large soft robot and rigid robot to move an object rigidly fixed between each end effector. Despite these advances, no prior work using multiple soft robotic manipulators shows forceful coordination through contact-rich manipulation of significant loads in all six \acp{DoF}.

\section{Material and Methods}

\subsection{Hardware Description}
\label{sec:hardware-description}
\begin{figure}[t]
    \centering
    \includegraphics[width=\columnwidth]{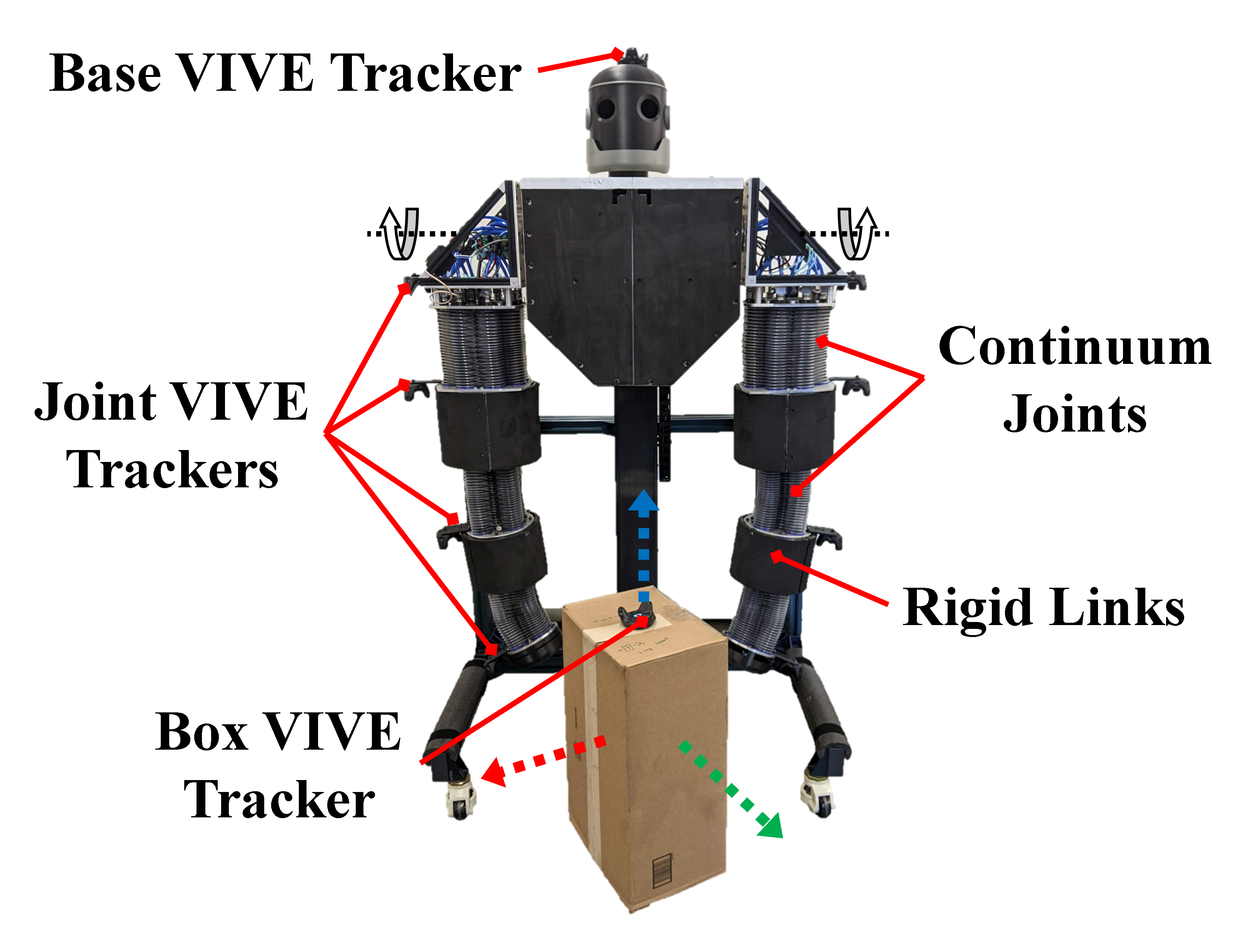}
    \caption{\blackout{Baloo}, a large-scale bimanual robotic manipulator with two pneumatically-driven robot arms mounted on a torso. Each arm can rotate forwards and be pinned into place. The external VIVE trackers for state estimation are labeled.}
    \label{fig:baloo-hw}
\end{figure}

We demonstrate whole-body grasping using \blackout{Baloo}, a large-scale hybrid rigid-soft robotic torso shown in Figure~\ref{fig:baloo-hw} and described in detail in \cite{johnson2024baloolargescalehybridsoft}. \blackout{Baloo} consists of two pneumatically actuated arms mounted on a rigid torso. Each arm consists of three soft continuum joints interconnected with rigid linkages. Each linkage contains valves, electronics, and tubing for PneuDrive \cite{pneudrive}, the embedded pressure control system. PneuDrive reports the four pressures associated with each joint. The entire torso is mounted on a linear actuator which allows the arms to reach the ground. The arms each rotate forwards as denoted by the white arrows in Figure~\ref{fig:baloo-hw} and are pinned at a specific angle. The black surfaces on the chest and arms of \blackout{Baloo} are sheets of neoprene foam to provide additional compliance during manipulation tasks. 

The soft continuum joints are actuated with four accordion-like pressure chambers which are arranged antagonistically in two pairs. The chambers deflate to atmospheric pressure (i.e., no vacuum), so they are only capable of pushing against each other to cause a bending torque. The stiffness of each joint can be modulated by increasing the mean pressure of each antagonistic pair. 

We parameterize each joint with two bending \acp{DoF}, using the \ac{CC} formulation presented in \cite{Allen_Rupert_Duggan_Hein_Albert_2020}. We do not include any twisting \acp{DoF} because the joints are very stiff in torsion. Each antagonistic pair of pressure chambers causes bending around a single \ac{DoF}. The configuration positions and velocities of the joints are estimated using VIVE motion trackers (see \cite{hyatt2018configuration} for similar estimation method) mounted to the rigid end plates of each joint, as shown in Figure~\ref{fig:baloo-hw}. 

The linear actuator is controlled with a stepper motor following time-optimal position trajectories with velocity, acceleration, and jerk constraints. We used the Ruckig Motion Planning library \cite{berscheid2021jerk} to plan these trajectories in real time. We measure both the generalized position and velocity of this \ac{DoF} through the stepper driver. 

For whole-arm manipulation tasks, we measure the pose (i.e., position and orientation) and twist (i.e., linear and angular velocity) of \blackout{Baloo's} base and the manipuland which, in our case, is a set of five boxes (see Table~\ref{tab:experiment_results}) using two additional VIVE trackers. The base frame pose/twist is necessary to map the box pose/twist to the `world frame', which is located on the ground directly beneath the linear actuator. One VIVE tracker is rigidly attached to each box directly above its geometric centroid. This tracker pose is transformed into the box centroid pose through a simple rigid-body transformation depending on the size of each box, illustrated by the dashed lines in Figure~\ref{fig:baloo-hw}. We use the resulting pose and twist of the box centroid relative to the `world frame' throughout this work.

\subsection{Simulation Description}
\label{sec:simulation-description}
To enable learned manipulation strategies, we model the \blackout{Baloo} soft robot in MuJoCo \cite{todorov2012mujoco}. The resulting simulation environment, \blackout{baloo}-mujoco-sim, is released open-source to facilitate reproducibility and future research. The following sections discuss the soft joints, their actuation mechanism, and the rigid linkages.

\subsubsection{Soft Continuum Joints}
\begin{figure}[t]
    \centering
    \subfloat[Constant curvature]{%
        \includegraphics[width=0.95\columnwidth]{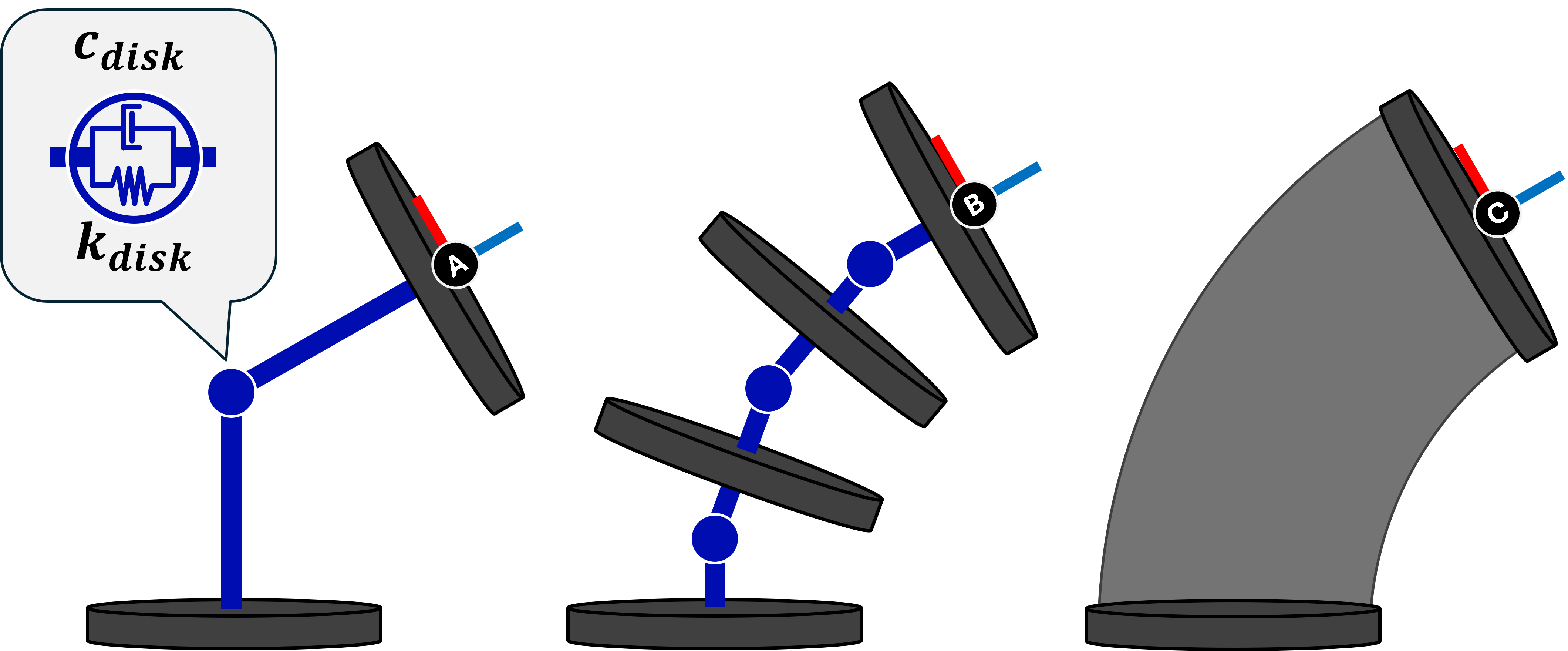}%
        \label{fig:CC_approx}%
    }
    \par
    \subfloat[Non-constant curvature]{%
        \includegraphics[width=0.85\columnwidth]{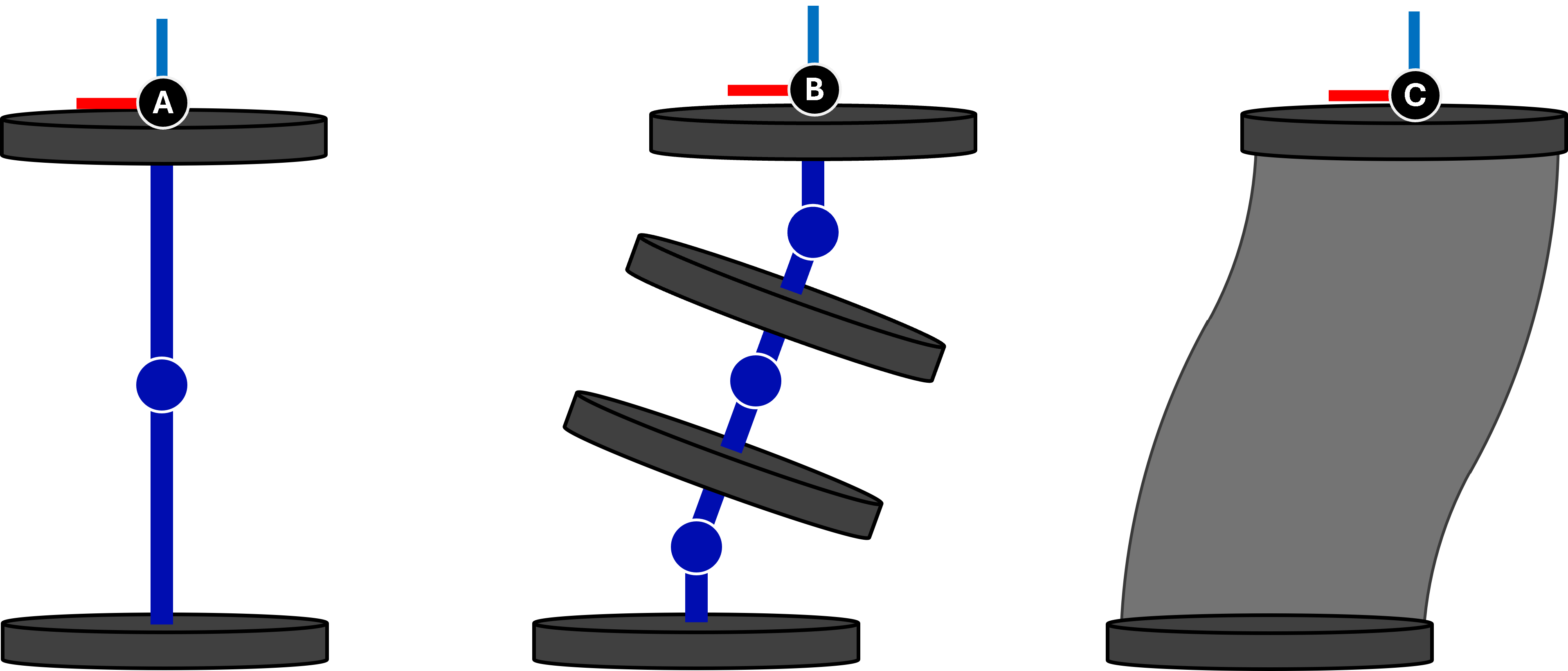}%
        \label{fig:nonCC_approx}%
    }
    \caption{\ac{UJ} approximations with $N=2$ and $N=4$ disks compared to a true continuum joint in constant curvature (a) and non-constant curvature (b). Each blue circle represents a \ac{UJ} with stiffness $k_{disk}$ and damping $c_{disk}$.}
    \label{fig:universal appox}
\end{figure}

We approximate the kinematics and dynamics of the soft continuum joints with a series of $N$ thin disks, interconnected with $N-1$ \acp{UJ} as shown in Figure~\ref{fig:universal appox}. The disks are spaced evenly along the length of the continuum joint and have a uniform mass distribution and standard moment of inertia of a solid cylinder. The overall mass of each continuum joint is evenly divided between each of the disks. 

\begin{figure}[t]
    \centering
    \includegraphics[width=0.7\columnwidth]{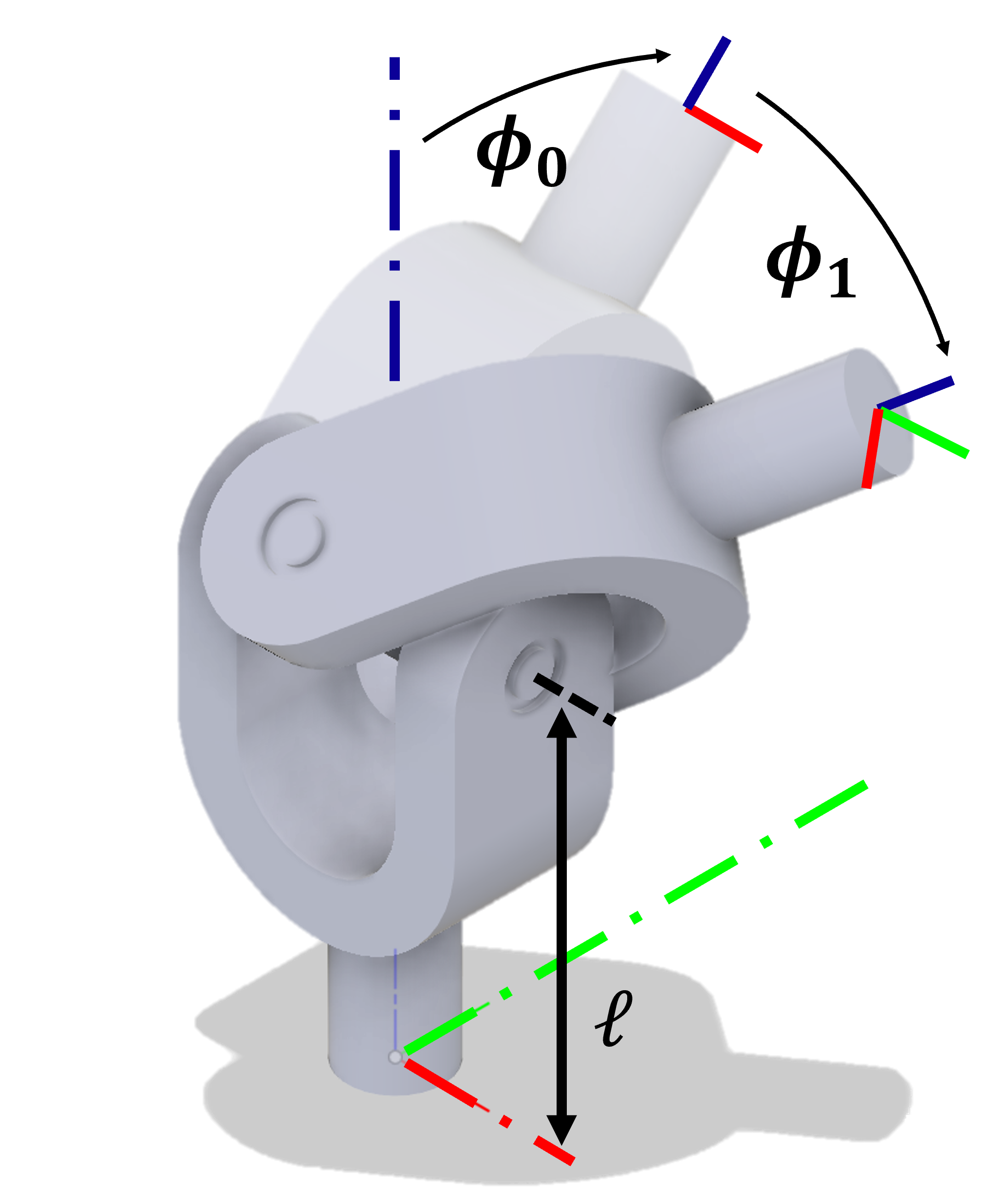}
    \caption{Illustration of a single \ac{UJ} with two configuration variables $\phi_0$ and $\phi_1$. Each segment is length $\ell$ for a total height of $2 \ell$. Equation~\ref{eq:universal-fk} defines the forward kinematics from the base to the tip of the joint. Note that although the visualization suggests possible linkage interference, the simulated joints in MuJoCo are ideal and do not experience collisions.}
    \label{fig:universal-joint-cad}
\end{figure}

A \ac{UJ} is connected to the centroid of each disk providing two \acp{DoF} between any pair of disks. The \ac{UJ} is modeled as an ideal \ac{UJ}, which is functionally equivalent to two co-located, perpendicular revolute joints, as shown in Figure~\ref{fig:universal-joint-cad}.

The forward kinematics of a single \ac{UJ} is

\begin{equation}
\label{eq:universal-fk}
\mathbf{G}(\phi_1, \phi_2) =
\begin{bmatrix}
c \phi_2 & 0 & s \phi_2 & \ell s\phi_2 \\
s\phi_1 s\phi_2 & c\phi_1 & -s\phi_1 c\phi_2 & -\ell s\phi_1 c\phi_2 \\
-c\phi_1 c\phi_2 & s\phi_1 & c\phi_1 c\phi_2 & \ell c\phi_1 c\phi_2 + \ell \\
0 & 0 & 0 & 1
\end{bmatrix}
\end{equation}

\begin{figure}[t]
    \centering
    \includegraphics[width=.8\columnwidth]{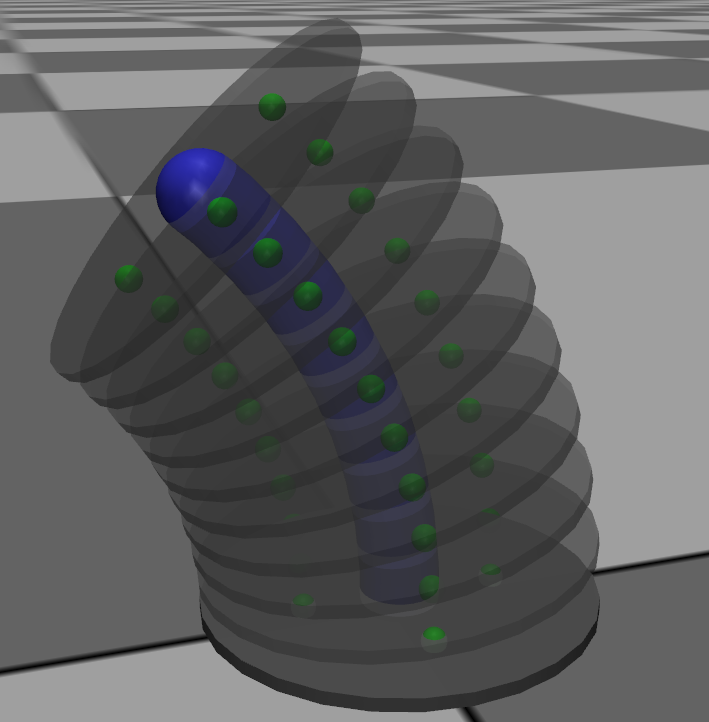}
    \caption{Semi-transparent view of the \ac{UJ} approximation implemented in MuJoCo with $N=10$ rigid disks. The \acp{UJ} are connected along the center of each disk, forming the blue curve (also shown in Figure~\ref{fig:universal appox}). The tendon actuators are constrained to pass through the green points and apply pneumatic forces in between each of the disks.}
    \label{fig:continuum-approx}
\end{figure}

By modeling several of these pairs together serially, we can approximate the continuum shape of the soft joints, as shown by the blue centerline in Figure~\ref{fig:universal appox}. Figure~\ref{fig:continuum-approx} shows this implemented in MuJoCo. The total transformation from the base to the tip after $n$ such segments is:

\begin{equation}
\label{eq:universal-fk-total}
T(\boldsymbol{\phi}) = \mathbf{G}^n(\boldsymbol{\phi}) = \prod_{i=0}^{n-1} \mathbf{G}(\phi_{2i}, \phi_{2i+1}) =
\begin{bmatrix}
\mathbf{R}(\boldsymbol{\phi}) & \mathbf{p}(\boldsymbol{\phi}) \\
\mathbf{0}_{1 \times 3} & 1
\end{bmatrix}
\end{equation}

\begin{equation}
\boldsymbol{\phi} = 
\begin{bmatrix}
\phi_0 & \phi_1 & \phi_2 & \cdots & \phi_{2n-2} & \phi_{2n-1}
\end{bmatrix}
\end{equation}

Note that we could also use ball joints in between each disk, but that would require additional torsional stiffness tuning and computation time due to the extra twisting \ac{DoF}. However, because the hardware joints we use are extremely stiff in torsion, we choose to use the \acp{UJ}. 

Internally, MuJoCo's simulation state includes the \ac{UJ} angles $\phi$ for each of the continuum joints but we abstract those values away the same way we do in hardware. By measuring the relative orientation between the base disk and tip disk, we can use the two \acp{DoF} \ac{CC} assumption to report two configuration positions and velocities for each joint. This estimation is implemented via a custom MuJoCo plugin. A planar case of this is illustrated in Figure~\ref{fig:CC_approx}. In Section~\ref{sec:results}, we will discuss the effects of this assumption, which can be violated during forceful interactions that cause the joint to shear as illustrated in Figure~\ref{fig:nonCC_approx}. 

We also assign stiffness and damping in universal-joint space. The overall stiffness ($K$) and damping ($C$) of a joint can be experimentally determined via passive deflection or dynamic testing. We assume that effective stiffness and damping of the entire joint is a result of equal springs ($k_{disk}$) and dampers ($c_{disk}$) connected in series between each disk (see Figure~\ref{fig:CC_approx}). The stiffness and damping coefficient in a joint with N disks is related to the overall stiffness and damping by 

\begin{equation}
    \frac{1}{K} = \frac{N-1}{k_{disk}}
\end{equation}

\noindent and

\begin{equation}
    \frac{1}{C} = \frac{N-1}{c_{disk}}.
\end{equation}

Joint limits can be implemented in three different ways: 

\begin{enumerate}
\item The first is a `hard' physical limit in \ac{UJ} space which limits the range of each \ac{UJ}. This can be implemented by dividing the joint's maximum bending angle into $N-1$ even parts (e.g. $90^\circ/9 = 10^\circ$ per \ac{UJ}). As we will discuss in Section~\ref{sec:results}, this assumption is perfect if the joint bends about a single \ac{DoF} but breaks down under non-planar motion. Thus this constraint by itself is not entirely sufficient to guarantee realistic joint limits.

\item The second method is through the ratio of actuation force to passive joint stiffness. Intuitively, this is a `soft' limit which requires tuning actuation gains and stiffness $K$ together to achieve the correct joint limit. In other words, at the correct joint limit the actuation force and the passive spring force should cancel each other out. This method requires more tuning but can result in more stable simulations at larger time steps.

\item The last method is the most physically realistic method to implement joint limits. This method relies on physical contacts between adjacent disks. The accordion-like pressure chambers of the soft robot \blackout{Baloo} compress until their grooves are physically touching. Any additional compression past this point is very difficult. By adjusting the thickness of adjacent disks in simulation, you can make contact between disks happen at different bending angles. The thickness of the disks simply needs to be adjusted so that contact between adjacent disks occurs at the joint limit. The largest downside of this method is that it requires contact simulation (which can be slow) and that the thickness needs to be individually tuned for different numbers of disks.
\end{enumerate}

In practice, we use combination of all three methods but emphasize the first and third. We start with the first method since it is easiest to measure, then check disk thicknesses to make sure the joint can bend the right amount passively, then tune the stiffness and actuation gains until the simulated pressures result in realistic bending angles. 

\subsubsection{Actuation}
We use the concept of `spatial tendons' in MuJoCo to create actuators which are constrained to pass through each of the green spheres shown in Figure~\ref{fig:continuum-approx}. There are four green spheres on each disk, representative of the four chambers that are used to actuate the joint. A unique feature of MuJoCo is its support for `stateful actuators' (i.e. actuators with internal dynamics), which is critical for simulating the pneumatics of a soft robot. By including the pressure state of the pneumatic chambers, the overall model becomes third order. We use this feature by attaching a cylinder actuator (i.e. pneumatic cylinder) to each spatial tendon. This has the effect of applying repulsive forces between each pair of green spheres, which in turn causes a bending torque to displace the \acp{UJ}. This is in contrast to other similar rigid-body approximations (e.g., SOMO \cite{graule2021somo}) of soft continuum joints where actuation is applied directly as a torque on each revolute joint along the spine. We designed our simulation this way because it more closely approximates the hardware which has actual pressure chambers that lie in the same locations. It also allows us to reason about and control actuation forces in pressure space instead of in an `imaginary' universal-joint space, which is generally abstracted away. In reality there are no well-defined joints about which to apply a torque on a continuum soft robot, making the mapping between commanded pressures and torques cumbersome without our proposed solution.

\subsubsection{Rigid Components}
The rigid links are modeled as cylinders with masses equal to the total mass of all of the hardware housed in the links on \blackout{Baloo}. We approximate the hardware internals with an inertia matrix of a solid cylinder aligned with its principal axes and a center of mass at its geometric centroid.

As mentioned previously, the entire torso is mounted on a linear actuator. We used the Ruckig Motion Planning library to plan smooth trajectories for the elevator on the hardware. We replicate this in MuJoCo with an additional custom plugin.

\subsection{\ac{RL} Environment Description}
\label{sec:env-description}

\subsubsection{Task Description}
Our objective is to teach the robot to lift objects from the ground using its whole body. Each episode is a maximum of 60 seconds long with the policy running at a rate of 20 Hz (i.e, 1200 steps). 

Learning-based approaches to robotic grasping typically use benchmark datasets of hundreds or thousands of objects to train on. These datasets are limited to small hand-held objects because manipulation is focused entirely on the end effector. No open-source benchmark dataset exists for the style of whole-body grasping on which we are focused. Hence, for the scope of this paper we choose to use boxes of varying shapes and weights. Boxes are a simple geometric primitive that is natively supported in MuJoCo. They are also inexpensive and easy to find in varying shapes for sim-to-real testing. A set of 1000 simulated training boxes were generated by Latin hypercube sampling parameters from the ranges shown in Table~\ref{tab:sim_params}. A box was randomly selected from this set for every new training episode. We chose the specific ranges in Table~\ref{tab:sim_params} heuristically to be small enough to fit within the robot's workspace, yet large enough to preclude simple end-effector grasping, as our focus is on whole-body grasping.

\begin{table}[t]
\centering
\caption{Range of values used to generate 1000 box objects for the training episodes.}
\label{tab:sim_params}
\begin{tabular}{@{}ll@{}}
\toprule
\textbf{Parameter}              & \textbf{Value}                               \\ \midrule
Manipuland Mass        &  0.5 - 10 kg \\
Manipuland Width        & 0.2 - 0.6 m \\
Manipuland Depth        &  0.2 - 0.6 m\\
Manipuland Height        &  0.5 - 1.25 m\\
\bottomrule
\end{tabular}
\end{table}

A given box was placed in front of the robot with a random initial position and orientation ( ranging from $\pm 0.1$ meters translation along x and rotation about z $R_z(\pm 60^\circ)$ respectively). Each episode outcome is categorized as a lift, a tip, or a slip. Tipping is defined as the object tilting more than 80 degrees from vertical. Success is defined as lifting the object 0.5 meters higher than its starting position. The episode terminates when the task is completed successfully or when the box tips over. The episode truncates (i.e., is categorized as a slip) when the time limit is reached since time is not part of the observation space \cite{pardo2022timelimitsreinforcementlearning, faramaDeepDive}. This usually happens because the box slipped out of reach, but did not tip over. 

\subsubsection{Observations}
The continuous measurement vector $o[t]$ at any given time step $t$ is described in Table~\ref{tab:observation_vector}. These values are measured directly from MuJoCo in the units shown and normalized to [-1,1] using the minimum and maximum bounds listed. The resulting normalized vector $\tilde{o}[t]$ is the observation vector used by the agent. 


\begin{table}[t]
\centering
\caption{Description of Measurement Vector $o[t]$}
\label{tab:observation_vector}
\resizebox{\columnwidth}{!}{%
\begin{tabular}{@{}lllll@{}}
\toprule
\textbf{Value}       & \textbf{Min}  & \textbf{Max} &\textbf{ Dim} & \textbf{Description (Unit)}                  \\ \midrule
$s$        & [.2,.2,.5]   & [.6,.6,1.25]   & 3   & Box Size (m)    \\
$p_o[t]$        & [-3,-3,0]   & [3,3,2]   & 3   & Box Position (m)    \\
$q[t]$        & -1   & 1   & 4   & Box Orientation (quat)    \\
$v[t]$        & -2   & 2   & 3   & Box Linear Velocity (m/s)        \\
$\omega[t]$        & -$\pi$   & $\pi$   & 3   & Box Angular Velocity (rad/s)        \\
$p_{\text{chest}}^{\text{box}}[t]$        & [0.2,0.2,0.5]   & [0.6,0.6,1.25]   & 3   & Chest to Box (m)       \\
$h[t]$           & -1.5 & 0   & 1   & Elevator Height (m)          \\
$\dot{h}[t]$      & -1   & 1   & 1   & Elevator Velocity (m/s)      \\
$q_l[t]$     & $-\pi$  & $\pi$  & 6   & Left Joint Angles (rad)      \\
$q_r[t]$    & $-\pi$  & $\pi$  & 6   & Right Joint Angles (rad)     \\
$\dot{q}_l[t]$  & $-2\pi$ & $2\pi$ & 6   & Left Joint Velocity (rad/s)  \\
$\dot{q}_r[t]$ & $-2\pi$ & $2\pi$ & 6   & Right Joint Velocity (rad/s) \\
$p_l[t]$     & 0    & 300 & 12  & Left Pressures (kPa)         \\
$p_r[t]$    & 0    & 300 & 12  & Right Pressures (kPa)        \\
$p_{l,cmd}[t]$     & 0    & 300 & 12  & Left Filtered Pressure Commands (kPa)         \\
$p_{r,cmd}[t]$     & 0    & 300 & 12  & Right Filtered Pressure Commands (kPa)              \\\bottomrule
\end{tabular}
}
\end{table}

\subsubsection{Actions}
The physical inputs to the system consist of 24 pressure commands and one height command. These values are continuous. We will refer to this vector as 

\begin{equation}
    u[t] = \Big(h_{des}[t], p_{l,des}[t], p_{r,des}[t] \Big) \in \mathcal{R}^{25}.
\end{equation}

\noindent $h_{des}[t] \in [-1.5,0]$ is the elevator height command in meters and $p_{i,des} \in [0,300]^{12}$ are the pressure commands in kPa where $i$ indicates the left ($i=l$) or right ($i=r$) arm.

In an effort to minimize the dimension of the action space and accelerate learning convergence, we parameterized $u[t]$ with a smaller continuous action vector given by 

\begin{equation}
\label{eq:a[t]}
    a[t] = [h_{des}, \Delta p_{l,des}, \Delta p_{r,des}] \in \mathcal{R}^{13}.
\end{equation}

\noindent where $h_{des} \in [-1,1] $ is the elevator height command and  $\Delta p_{i, des} \in [-150,150]^6 $ is a 6D differential pressure vector where $i$ indicates the left ($i=l$) or right ($i=r$) arm.  $\Delta p_{i, des}$ is transformed into 12 antagonistic pressure commands defined as $p_{i,des}$, centered at 150 kPa using the following mapping:

\begin{equation}
    p_{i,des} = 150 + (I_6 \otimes A)\Delta p_{i,\text{des}}
\end{equation}

\noindent where
\begin{equation}
    A = 
    \begin{bmatrix}
        1 \\
        -1\\
    \end{bmatrix}.
\end{equation}

\noindent and the $\otimes$ operator denotes the Kronecker product. During rollouts, the output of the learned policy is a vector of normalized actions $\tilde{a}[t] \in [-1,1]^{13}$ which is low pass filtered and then unnormalized to $a[t]$ using the bounds described above. Then $a[t]$ is mapped to $u[t]$ which is sent to the robot.

\subsubsection{Reward Engineering for Whole-Body Grasping}
\label{sec:reward-engineering}
We used a combination of sparse task completion/failure rewards and a shaped reward. The sparse task reward term is 

\begin{equation}
\label{r_task}
    r_{\mathrm{task}}  =   
\begin{cases}
      -2, & \text{if box tipped} \\
      10, & \text{if box height $>$ initial height + 0.5 m.}
\end{cases}
\end{equation}

After testing various reward terms, we found it challenging to guide the robot effectively towards stable grasps. This difficulty arises partly from differences in system dynamics (with the elevator moving much slower than the arms for safety), as well as the tendency of learning algorithms to fall into undesirable local minima (i.e., reward hacking), compounded by the challenges of exploring the high-dimensional action space.

Consequently, we opted to encode our preferences using a single term based on a motion primitive. This proved to be more stable and better at avoiding local minima compared to several reward terms encouraging stability, proximity, contact, safety, etc. This idea is similar to \ac{IL} approaches like behavior cloning or inverse \ac{RL} \cite{hussein2017imitation} but is distinct since it does not try to learn from demonstrations directly. Instead, exploration is indirectly guided towards the motion primitive by shaping the reward term \cite{Zhang_Barreiros_Onol_2023, 8463162, pertsch2021skild}. 

This approach reduces the burden on demonstrations to be optimal (or even successful) since they only exist to nudge the agent towards promising regions of the state space \cite{brys2015reinforcement}. Although this method requires access to an expert action vector instead of just standard state-action-reward tuples, defining such a policy was straightforward in our case. The motion primitive allowed us to encode several desirable behaviors in a single term implicitly: (i) avoiding collisions with the ground, (ii) grasping with both arms, and (iii) delaying the grasp until the torso is sufficiently close to the object.

\begin{figure}[t]
    \centering
    \subfloat[]{%
        \includegraphics[width=0.24\columnwidth]{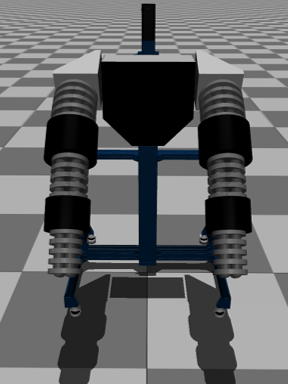}%
        \label{fig:zero_waypoint}%
    }
    \hfill
    \subfloat[]{%
        \includegraphics[width=0.24\columnwidth]{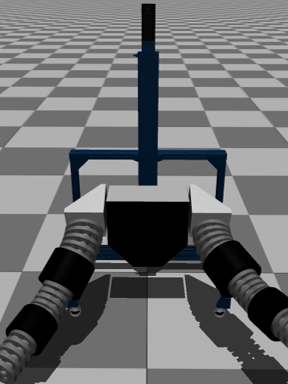}%
        \label{fig:approach_waypoint}%
    }
    \hfill
    \subfloat[]{%
        \includegraphics[width=0.24\columnwidth]{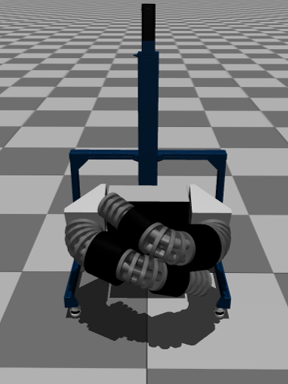}%
        \label{fig:grasp_waypoint}%
    }
    \hfill
    \subfloat[]{%
        \includegraphics[width=0.24\columnwidth]{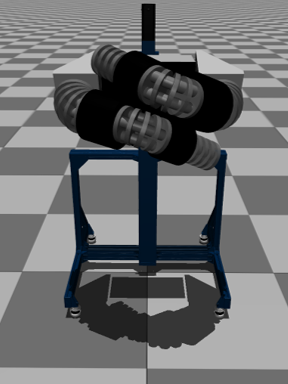}%
        \label{fig:lift_waypoint}%
    }
    \caption{Robot pose resulting from each waypoint described in Algorithm~\ref{alg:open-loop-hugger}. The motion primitive (i.e., trajectory) used to guide exploration is a linear interpolation between these waypoints.}
    \label{fig:waypoints}
\end{figure}

The motion primitive is based on two manually chosen waypoints, referred to as $\Delta p_{\text{approach}}$ and $\Delta p_{\text{grasp}}$. Each waypoint is a set of differential pressures which cause the arms to reach and grasp and was chosen in a few minutes of manual experimentation in simulation. The resulting robot pose of each waypoint is shown in Figure~\ref{fig:waypoints}. We generate a trajectory based on this motion primitive for an entire episode with a piecewise linear interpolation between each waypoint, as shown in Algorithm~\ref{alg:open-loop-hugger}.

The shaped motion-primitive term is a quadratic exponential

\begin{equation}
    r_{guide} = 0.1 e^{-0.5 ||\tilde{a} - \tilde{a}^*||^2}
    \label{eq:r_guide}
\end{equation}

\noindent where $\tilde{a} \sim \pi_\theta(\cdot|s)$ is the action predicted by the current policy and $a^*$ is the action given by the motion primitive described in Algorithm~\ref{alg:open-loop-hugger}. Thus the total reward at a given timestep is 

\begin{equation}
\label{eq:full-guided-reward}
    r[t] = r_{\mathrm{task}} + r_{\mathrm{guide}}.
\end{equation}

\begin{algorithm}
\caption{Waypoint-based Motion Primitive used to construct the reward term in eq.~\eqref{eq:r_guide}.}
\label{alg:open-loop-hugger}
\begin{algorithmic}[]
\Require elevator height $h[t]$, $\Delta p_{\text{approach}}$, and $\Delta p_{\text{grasp}}$.
\State $n \gets 0$, $N \gets 50$ 

\State $\text{state} \gets \text{APPROACH}$ \Comment{see Fig. \ref{fig:zero_waypoint}}

\If{$\text{state} = \text{APPROACH}$}
    \If{$n < N$}
        \State $h_{des} \gets -0.9$
        \State $\Delta p_{l,des} \gets \frac{n}{N} \Delta p_{l,\text{approach}}$
        \State $\Delta p_{r,des} \gets \frac{n}{N} \Delta p_{r,\text{approach}}$
        \State $n \gets n + 1$
    \ElsIf{$h[t] \approx -0.9$} \Comment{see Fig. \ref{fig:approach_waypoint}}
        \State $n \gets 0$
        \State $\text{state} \gets \text{GRASP}$ 
    \EndIf
\ElsIf{$\text{state} = \text{GRASP}$}
    \State $\Delta p_{l,des} \gets (1-\frac{n}{N}) \Delta p_{l,\text{approach}} + \frac{n}{N} \Delta p_{l,\text{grasp}}$
    \State $\Delta p_{r,des} \gets (1-\frac{n}{N}) \Delta p_{r,\text{approach}} + \frac{n}{N} \Delta p_{r,\text{grasp}}$

    \State $n \gets n + 1$

    \If{$n = N$} \Comment{see Fig. \ref{fig:grasp_waypoint}}
        \State $\text{state = LIFT}$ 
    \EndIf
\ElsIf{$\text{state} = \text{LIFT}$}
    \State $h_{des} \gets 0$ \Comment{see Fig. \ref{fig:lift_waypoint}}
\EndIf
\State $a[t] = [h_{des}, \Delta p_{l,des}, \Delta p_{r,des}]$
\State $\tilde{a}[t] = \text{normalize}(a[t])$\\
\Return $\tilde{a}[t]$
\end{algorithmic}
\end{algorithm}

\subsubsection{Learning Algorithm}
We use the Proximal Policy Optimization (PPO) implementation available in Stable-Baselines3 (SB3) \cite{stable-baselines3}. As a brief overview, PPO jointly optimizes an actor (stochastic policy network $\pi_\theta(a|s)$) and a critic (a Value Function $V_\theta(s)$). The combined objective function is comprised of three terms

\begin{equation}
L_t(\theta)=\hat{\mathbb{E}}_t\left[L_t^{C L I P}(\theta)-c_1 L_t^{V F}(\theta)+c_2 S\left[\pi_\theta\right]\left(s_t\right)\right]
\end{equation}

\noindent where $\hat{\mathbb{E}}[\cdots]$ is an estimated expectation over a finite batch of trajectories, $L_{CLIP}$ is the clipped policy loss, $L_{VF}$ is the value function loss, and $S[\pi_\theta](s_t)$ is the entropy bonus. Clipping the policy loss prevents excessively large policy updates to stabilize training. Minimizing $L_{VF}$ improves credit assignment and training efficiency. $S[\pi_\theta](s_t)$ is a mechanism to encourage exploration by increasing the entropy of the policy distribution $\pi_\theta(a|s)$. In our network architecture, the policy loss gradients are only affected by $L_{CLIP}$ and $S[\pi_\theta](s_t)$ while the value function loss gradients are affected by $L_{VF}$, since there are no learnable parameters shared between the policy and value networks. Both networks are feed-forward with two hidden layers. 
Table~\ref{tab:ppo-hyperparams} reports the PPO hyperparameters used in this work. We identified these empirically.

\begin{table}[t]
\centering
\caption{PPO Training Hyperparameters}
\label{tab:ppo-hyperparams}
\begin{tabular}{@{}ll@{}}
\toprule
\textbf{Hyperparameter}      & \textbf{Value }       \\ \midrule
n\_steps            & 8192         \\
n\_envs             & 16           \\
gamma               & .995         \\
batch\_size         & 512          \\
learning rate       & 3e-4 to 1e-5 \\
ent\_coeff          & 0.0          \\
value network arch  & 512, 512     \\
policy network arch & 512, 512      \\
vf\_coeff           & 0.5          \\
max\_grad\_norm     & 0.5          \\
clip\_range         & 0.2          \\
gae\_lambda         & 0.95         \\
log\_std\_init      & 0.0          \\ 

\bottomrule
\end{tabular}
\end{table}


\subsection{Experimental Protocols}
\label{sec:experiment-protocol}

This section summarizes the methods used in each experiment. First, we examined the speed vs. accuracy tradeoffs of our simulation framework. Then, we evaluated the trained policy in simulation. Finally, we performed a zero-shot, sim-to-real transfer to hardware.

\subsubsection{Simulation Speed vs Accuracy}
\label{sec:speedvaccuracy}
To evaluate the speed and accuracy of our continuum joint approximation, we performed three experiments:

\textbf{Single-threaded \ac{RTF}:} We evaluated the nominal performance of the simulation on a single thread using the \texttt{testspeed} binary included in MuJoCo. Each test is an average of 10000 simulation time steps. For each data point, we varied the time step duration from .1 ms to 10 ms and the number of disks $N$ from 2 to 64. The resulting \ac{RTF} is the average ratio of simulated time to wall clock time. For example, a \ac{RTF} of 5 means that the simulation can simulate five seconds in one real-world second.

Note that we report single-threaded speeds because it accurately captures performance independent of machine configuration. Multi-threading does boost the effective \ac{RTF} by parallelizing each simulation, but we observed diminishing returns past a certain number of threads. This is mainly due to hardware configuration such as core count and memory bandwidth. The optimal number of threads will vary, depending on the computer configuration. 

\textbf{\ac{CC} vs \ac{UJ} Approximation:} The objective of this experiment was to quantify how well the kinematics (Eq~\ref{eq:universal-fk-total}) of a series of $N \in \{2,4,8,16,32, 64\}$ disks connected by $N-1$ \acp{UJ} can approximate the kinematics of the well-known \ac{CC} model \cite{Allen_Rupert_Duggan_Hein_Albert_2020} in terms of tip pose. Because we used the \ac{CC} kinematic model to estimate joint configurations on hardware, we also used it in simulation to estimate $q$ and $\dot{q}$ in Table~\ref{tab:observation_vector}.  Thus this experiment reveals any modeling discrepancies between the \ac{UJ} approximation and our current estimation setup.

When localized non-\ac{CC} deformation occurs during whole-arm manipulation, there are actual errors between the true pose of the joint and the \ac{CC} model's estimate. This experiment does not consider the error between `ground truth' (i.e., the actual configuration of the joint in hardware) and the \ac{CC} model, only modeling discrepancies between the \ac{CC} model and the \ac{UJ} approximation. The \textit{non-\ac{CC} vs \ac{UJ} Approximation} experiment will examine these additional errors.

We computed the predicted tip pose using the \ac{CC} model and the \ac{UJ} model on an equal spaced samplingof \ac{CC} configuration variables ranging from -2.1 to 2.1 radians, which roughly corresponds to the maximum bending angle of the joints.

Given a \ac{CC} configuration $q$, it is straightforward to compute the predicted tip pose using the \ac{CC} model. But in order to compute the predicted tip pose using the \ac{UJ} model (Eq~\ref{eq:universal-fk-total}), we need to map $q$ to a suitable set of \ac{UJ} angles $\phi \in \mathcal{R}^{2n-1}$. To choose from the infinite set of possible solutions, we solved an inverse kinematics problem to find a $\phi$ that minimized the difference in tip pose between the \ac{CC} and \ac{UJ} predictions. We assume that any residual error that could not be eliminated during optimization is due to a fundamental kinematic mismatch between the \ac{UJ} approximation and \ac{CC} models. 

Given $q$ and $N$, the inverse kinematics problem is defined as

\begin{equation}
\label{eq:fk_min}
\begin{aligned}
\min_\phi \quad &\mathcal{E}^{pos}(\phi, q) + \mathcal{E}^{ori}(\phi, q) + \mathcal{E}^{reg}(\phi) \\
\text{subject to} \quad & p_{\text{CC}}, R_{\text{CC}} = \text{FK}_{\text{CC}}(q)
\end{aligned}
\end{equation}

\noindent where $\text{FK}_\text{CC}$ is the \ac{CC} forward kinematic model from \cite{Allen_Rupert_Duggan_Hein_Albert_2020}, and $p_{cc} \in  \mathbb{R}^3$ is the resulting position vector and $R_{cc} \in \mathbb{R}^{3x3}$ is the resulting rotation matrix. The terms of the objective function are defined as

\begin{align}
    \mathcal{E}^{pos}(\phi, q) &= \left\| p(\phi) - p_{\text{CC}}(q) \right\|^2\label{eq:first-term}\\
    \mathcal{E}^{ori}(\phi, q) &= \lambda^2 \left\| \log \left( R(\phi)^\top R_{\text{CC}}(q) \right)^\vee \right\|^2\label{eq:second-term}\\
    \mathcal{E}^{reg}(\phi) &= \alpha \|\phi\|^2.\label{eq:third-term}
\end{align}

$p(\phi)$ is the tip position vector and $R(\phi)$ is the tip orientation matrix given in Equation~\ref{eq:universal-fk-total}. $\text{log}(\cdot)$ is the matrix logarithm for SO(3) and $(\cdot)^\vee$ maps a skew-symmetric matrix to a 3-vector. Together they map the rotation matrix to an axis-angle representation. $\lambda$ is a normalizing term to weight position and orientation equally. Equation~\ref{eq:third-term} is a regularizer to encourage the optimizer to `center' the \acp{UJ}. 

\textbf{Non-\ac{CC} vs \ac{UJ} Approximation:} The objective of this experiment was to quantify how well the \ac{UJ} approximation can capture non-constant curvature shapes. As noted previously, though the \ac{CC} model has been used successfully in prior work, it breaks down in the presence of external forces that cause localized deformation which we expect to be common during whole-body manipulation (see Figure~\ref{fig:nonCC_approx}). Ideally, we would compare the \ac{UJ} simulation directly to hardware under different loading conditions, but measuring the true shape of a continuum joint in hardware is an active area of research and beyond the scope of this work. Instead, we treat a slow, finely discretized \ac{UJ} model with $N=64$ disks as ground truth, assuming it closely approximates the difficult-to-measure, true hardware continuum. We ran full manipulation episodes with coarser models ($N=2 \rightarrow 32$) to quantify how well the \ac{UJ} approximation captures the continuum behavior relative to the high-resolution baseline.

\subsubsection{Policy Training}
We compare our guided learning approach against a shaped reward baseline consisting of: (1) chest proximity, (2) number of contacts, and (3) box lifting height. The shaped reward is:

\begin{equation}
\label{eq:full-shaped-reward}
    r = r_{\text{task}} + r_{\text{approach}} + r_{\text{grasp}} + r_{\text{height}},
\end{equation}

with

\begin{align}
r_{\text{approach}} &= 0.1 \cdot e^{-4 ||p_{\text{chest}}^{\text{box}}||^2} \\    
r_{\text{grasp}} &= 
\begin{cases}
0.1 \cdot N_{\text{contacts}}, & \text{if } |(p_{\text{chest}}^{\text{box}})_z| < 0.35 \\
0, & \text{otherwise}
\end{cases} \\
r_{\text{height}} &= \max\big(0, (p_o[t])_z - (p_o[0])_z\big)
\end{align}

where $p_{\text{chest}}^{\text{box}} = p_{\text{chest}} - p_{\text{box}}$ and $p_o[t]$ is cartesian position of the box at time $t$ (see Table~\ref{tab:observation_vector}). $r_{\text{task}}$ matches the guided reward’s task term (Eq. \ref{r_task}).

This reward draws from preliminary work inspired by \cite{alessi2024pushing,Vulin_Christen_Stevšić_Hilliges_2021,agabiti2023whole,Raza}, which use shaped rewards to guide multi-step tasks such as approach, grasp, and manipulation. It serves as a natural baseline for this RL problem.

\subsubsection{Policy Evaluation in Simulation}
We conducted a series of simulation experiments to systematically evaluate the performance and robustness of the learned policy compared to the baseline on 1000 randomly generated boxes. The evaluation focused on the aspects in each section below. 

\textbf{Object-wise Outcome Analysis:} We characterized the behavior of the learned policy on 1000 randomly generated boxes, collecting statistics on the outcome of each episode.

To analyze trends in failure cases, we computed the Spearman correlation coefficient $\rho \in [-1, 1]$ between the policy's outcomes and the experimental variables (e.g., size, mass, success rate, etc). The Spearman correlation coefficient quantifies how well the relationship between two variables can be described using a monotonic function, even if it is nonlinear. The sign of the coefficient indicates if the monotonic function is positive or negative. 

For example, if there is a strong positive correlation between success rate and box height, this indicates that 1) taller boxes are easier to lift and that 2) shorter boxes are more difficult to lift. In fact, if there are significant non-zero correlations between success rate and other experimental inputs, the success rate is dependent on those inputs. If the success rate was affected by inputs, we could conclude that the policy has a lower degree of generalization on a subset of objects. In summary, a policy that generalizes well across all inputs will have success rates with low correlations across all inputs. 

\textbf{Motion Primitive vs Policy:}
The initial objective of this comparison was to quantify the performance improvement of the learned policy relative to the baseline Primitive policy in an ideal, unperturbed environment.

To achieve this, we executed trials for both the open-loop motion primitive and the Learned policy on a set of 1000 distinct test boxes, each starting from an identical initial pose. We recorded the final outcome for each trial (Success, Slip, or Tip). To visualize the improvements and regressions introduced by learning, we compiled these results into a transition matrix, which directly compares the outcome from the motion primitive with the outcome from the learned policy for each object. 

Given the deterministic nature of the simulation and policy evaluation, any differences in outcome must originate from differences in the applied action trajectories.

\textbf{Effect of External Perturbation on Outcomes:}
To evaluate the robustness of the learned policy and isolate its reactive capabilities, we applied external perturbations to the box during the grasp trial. 

The perturbations were 100 N downward forces at the box's centroid in one-second intervals (on for one second, off for one second). We chose this pattern because it was enough to consistently challenge the grasp stability without making success impossible. We conducted 1000 trials for both the motion primitive (which cannot react to the perturbations) and the learned policy under these conditions.

As with the non-perturbed trials, we compiled the results into a paired outcome matrix. By comparing these outcomes to the unperturbed results, we can examine the contribution of the policy's reactive components, since the open-loop primitive cannot respond to the disturbance. This comparison isolates the value of the learned, closed-loop behavior in maintaining grasp stability under stress.

\textbf{Learned Response to Perturbation:}
To investigate how the characteristics of the policy's corrective action correlate with different grasp outcomes, we performed an analysis of the corrective action magnitudes over time.

While the motion primitive is open-loop and cannot alter its actions, the learned policy can. To understand the mechanism of its reaction, we calculated the `corrective action' for the learned policy at each timestep $t$. This metric is defined as the L2 norm of the difference between the action vectors chosen by the policy with and without the 100 N downward perturbation $||\Delta a[t]|| = ||a_{\text{unperturb}}[t] - a_{\text{perturb}}[t]||$. To produce a standardized and comparable measure of effort, we normalized the corrective action by the maximum possible L2 norm of the 13-dimensional action vector $a$ (which is $\sqrt{13} \approx 3.606$), with the final metric reported as a percentage. This calculation allows us to isolate and quantify the reactive component of the learned policy's behavior.

Our outcome analysis focused on a fixed 100-step (5-second) window immediately following the initial onset of the perturbation. For each trial, we calculated the instantaneous corrective action, $||\Delta a[t]||$, at each timestep $t$ within this window. To identify patterns associated with outcome transitions, we grouped the trials based on their outcome transitions (e.g., Slip $\rightarrow$ Tip, Success $\rightarrow$ Success) and calculated summary statistics (mean, median, and max) over the entire 100-step window of the normalized corrective action across each group. 

This analysis reveals which corrective action characteristics are associated with successful grasp recovery versus those that led to failure.

\subsubsection{Zero-shot Sim-to-Real Transfer}
To evaluate the learned policy on hardware, we use a set of five cardboard boxes of varying shapes and masses reported in Table~\ref{tab:experiment_results}, which are within the training distribution in simulation (see Table~\ref{tab:sim_params}). We filled each box with different weights to span the 0.5 - 10 kg range in Table~\ref{tab:sim_params}. However, we note one substantial discrepancy between simulated and real boxes. In simulation, the boxes are assumed to have a center of mass at the centroid of the object and a uniform distribution and inertia. In reality, this is not true with the added weight laying at the bottom of otherwise empty boxes. The policy is never explicitly trained on this situation. Despite this gap, we did not fine-tune the learned policy on hardware. The policy is transferred zero-shot for all experiments. We conducted five trials for each box, totaling 25 trials. Similar to simulation, we recorded the outcome of each trial as success, tip, or slip. This experiment also serves to validate our proposed simulation method described in Section~\ref{sec:simulation-description} as being accurate enough to enable learning-based control methods for real soft-continuum-robot hardware.

\section{Results and Discussion}
\label{sec:results}
\subsection{Simulation Speed vs Accuracy}
\begin{figure}[t]
    \centering
    \includegraphics[width=\columnwidth]{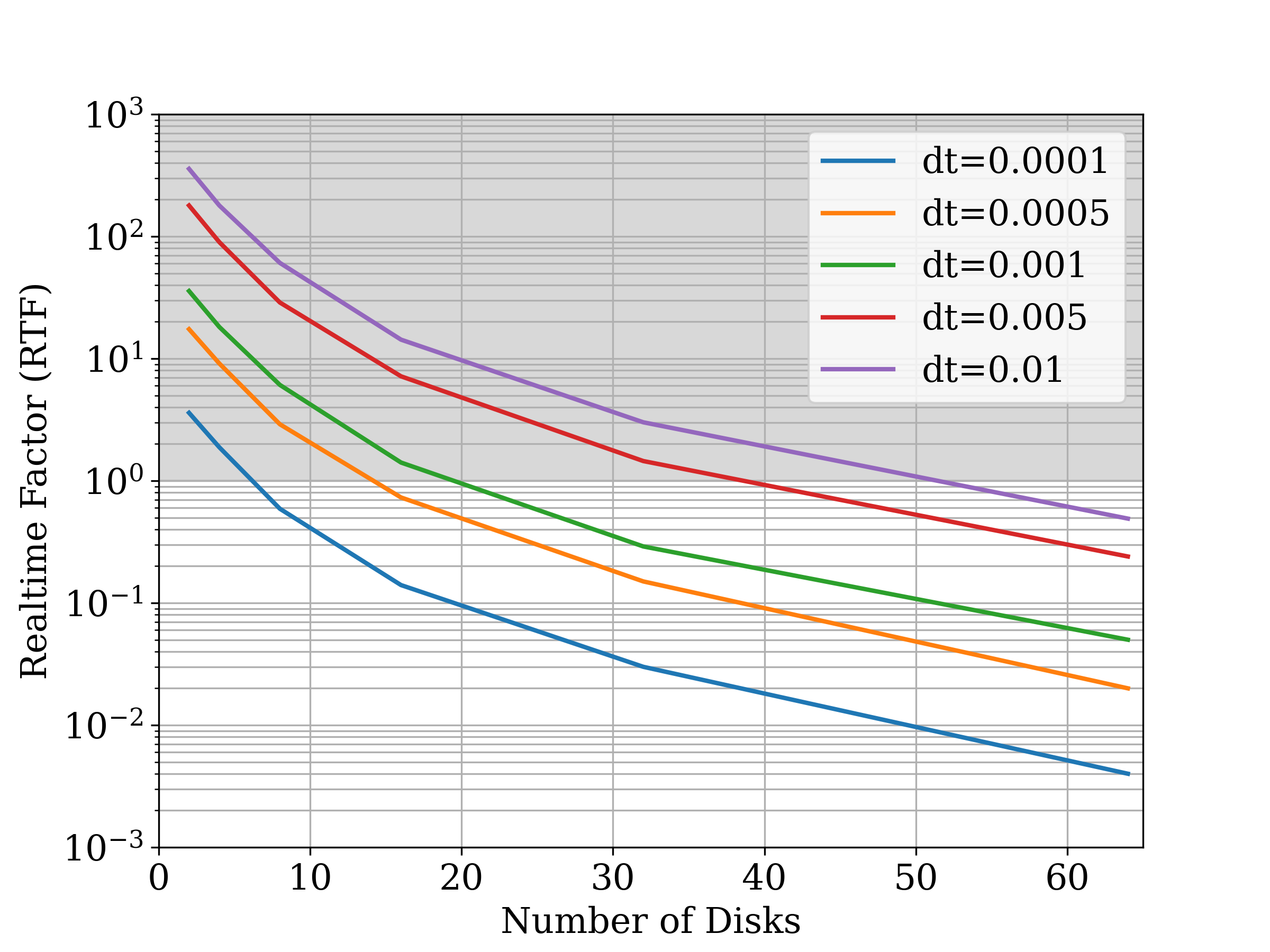}
    \caption{Results of \textit{Experiment 1}: Single-threaded real time factor as a function of simulation timestep and number of disks. Shaded region indicates real-time capability.}
    \label{fig:realtime-factor}
\end{figure}

\textbf{Single-threaded \ac{RTF}:} Figure~\ref{fig:realtime-factor} shows the single-threaded \ac{RTF} as a function of simulation time step and the number of disks. 
For the results in the following sections, we used a simulation timestep of 5 ms and 5 disks, resulting in a single-threaded \ac{RTF} of 65x. Fewer disks (i.e., $N < 5$) introduced too much error and made the lifting task difficult to accomplish. Larger time steps (e.g., .01 s) often worked well, but occasionally caused numerical problems when there were many contact constraints to solve. Since the policy runs at 20 Hz, its output (i.e., a vector of commanded pressures) is applied for 10 consecutive simulation steps.
We conducted our training runs on an AMD Ryzen 9 5900 CPU with 12 cores and 24 threads. With parallelization on 16 threads as shown in Figure~\ref{fig:baloo-rl}, we achieved an effective \ac{RTF} of approximately 550x.

\begin{figure*}[]
    \centering
    \subfloat[]{%
        \includegraphics[width=\columnwidth]{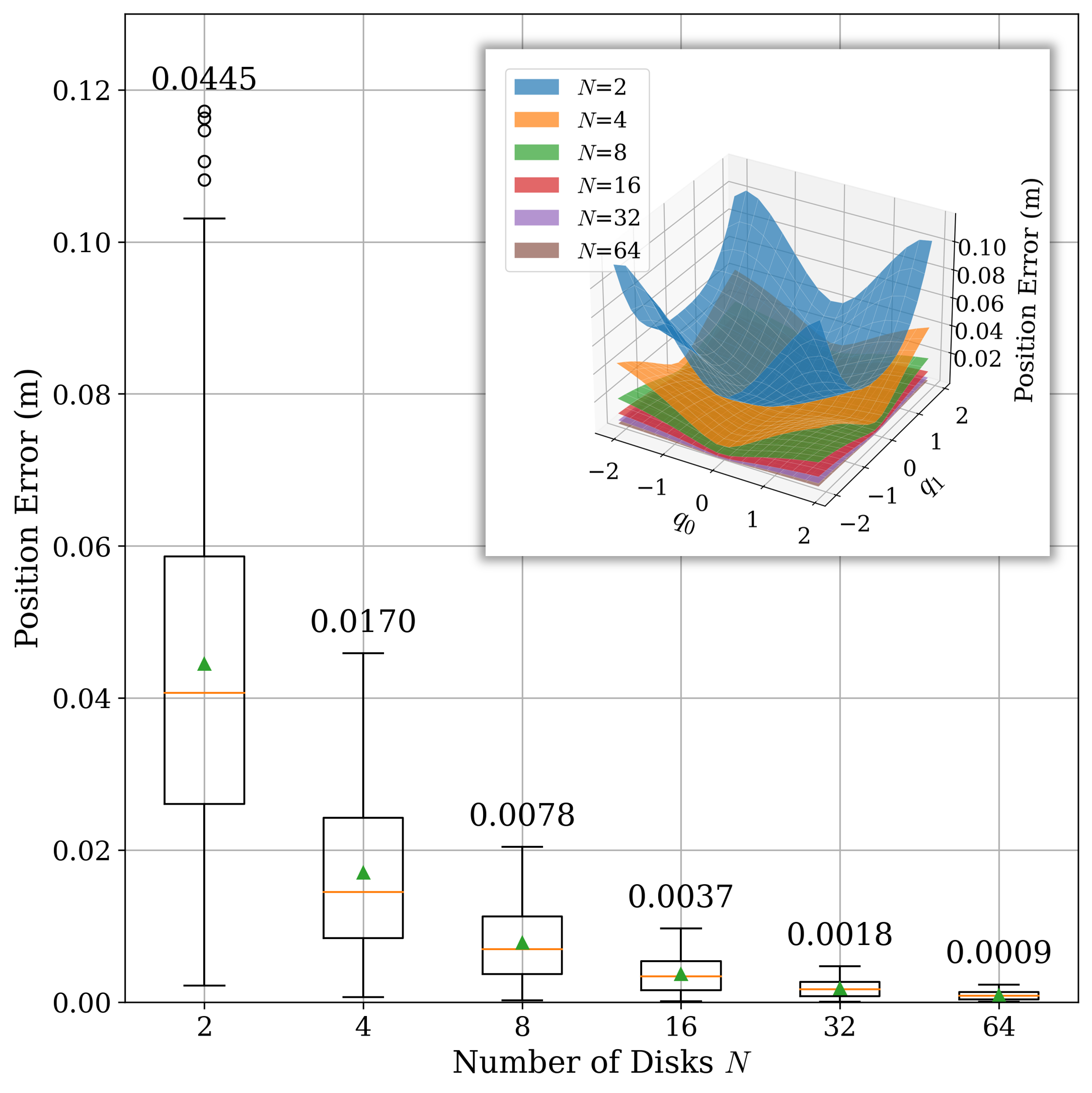}%
        \label{fig:tip_pos_error}%
    }
    \hfill
    \subfloat[]{%
        \includegraphics[width=\columnwidth]{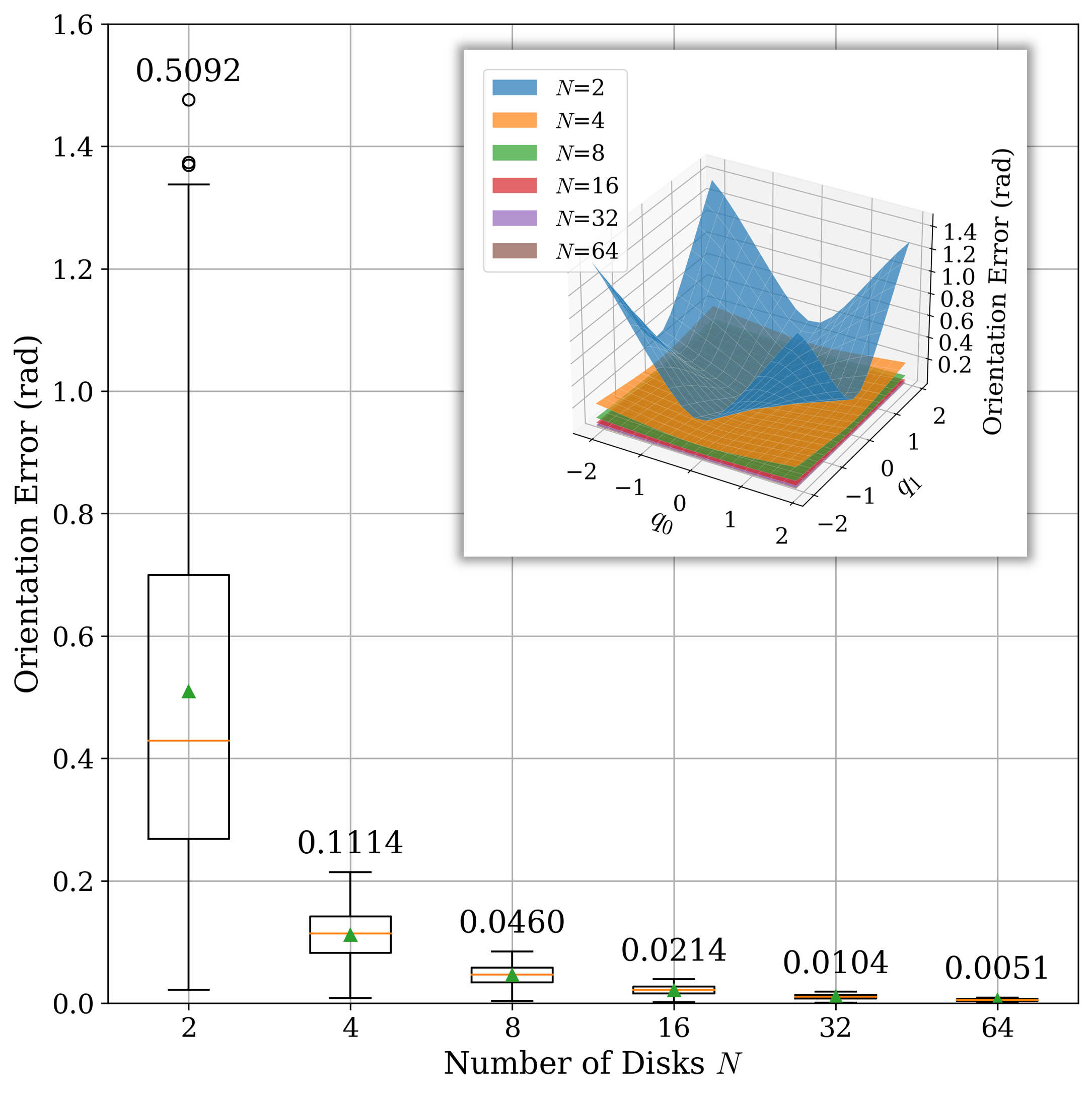}
        \label{fig:tip_ori_error}%
    }
    \caption{Results of \textit{{\ac{CC} vs \ac{UJ} Approximation}}: Tip position (a) and orientation (b) errors as a box plot and surface. Each box plot aggregates the data from one entire surface. The green triangles are means over the entire surface, and whose values are displayed above each box plot.}
    \label{fig:tip_errors}
\end{figure*}

\textbf{\ac{CC} vs \ac{UJ} Approximation:} Figure~\ref{fig:tip_errors} illustrates the residual errors in tip pose between the \ac{CC} and \ac{UJ} models, as a function of joint \ac{CC} configurations and numbers of disks (i.e., the pose error between point A/B and point C in Figure~\ref{fig:CC_approx}). Figure~\ref{fig:tip_pos_error} shows the tip position error on the vertical axis as a function of \ac{CC} configuration variables $q = [q_0, q_1]$ with the same data as a box plot for easier numerical comparison. Similarly, Figure~\ref{fig:tip_ori_error} shows the tip orientation error. Recall from Section~\ref{sec:speedvaccuracy} that we assume these errors are due to a fundamental kinematic mismatch between each model, as the \ac{UJ} angles have been chosen to minimize the tip pose error (Eq.~\ref{eq:fk_min}). This is a measure of how well \ac{CC} and \ac{UJ} approximation agree with one another in terms of position and orientation of the tip.
Figure~\ref{fig:tip_errors} reveals a few key trends. The tip pose error is always a function of configuration, regardless of the number of disks, but Figures \ref{fig:tip_pos_error} and \ref{fig:tip_ori_error} show that the error surface flattens out quickly. All of the errors exponentially approach zero as $N$ increases. 
The \ac{UJ} approximation clearly performs poorly with larger bend angles, resulting in the sharp surface inclines and the upward skewed box plots. To illustrate the skewness of the data, we also plot the averages as green triangles and display their values above each box plot. Until $N$ is large enough, the averages are higher than the medians plotted as orange horizontal lines.

\begin{figure*}[]
    \centering
    \subfloat[]{%
        \includegraphics[width=\columnwidth]{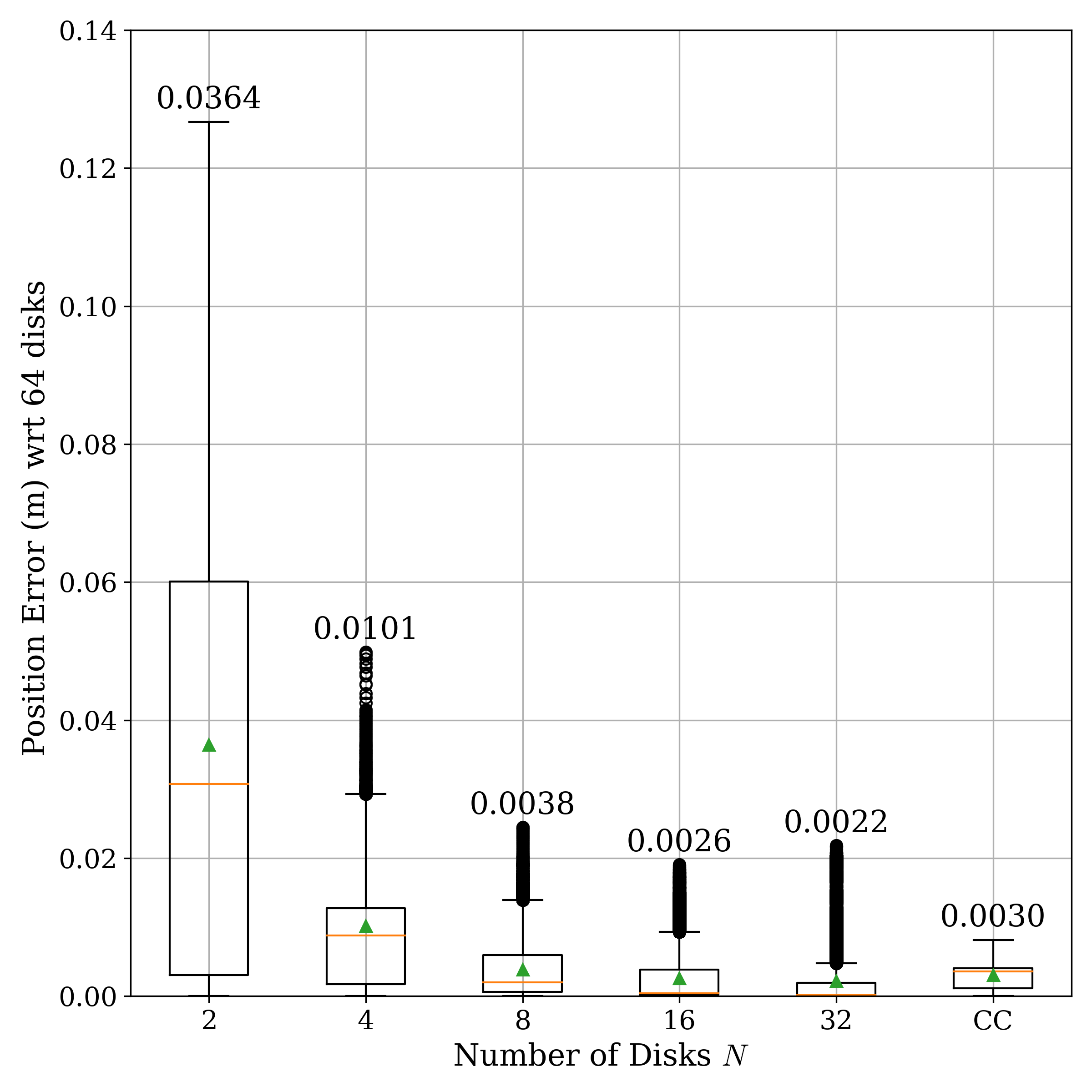}%
        \label{fig:tip_position_error_64disks}%
    }
    \hfill
    \subfloat[]{%
        \includegraphics[width=\columnwidth]{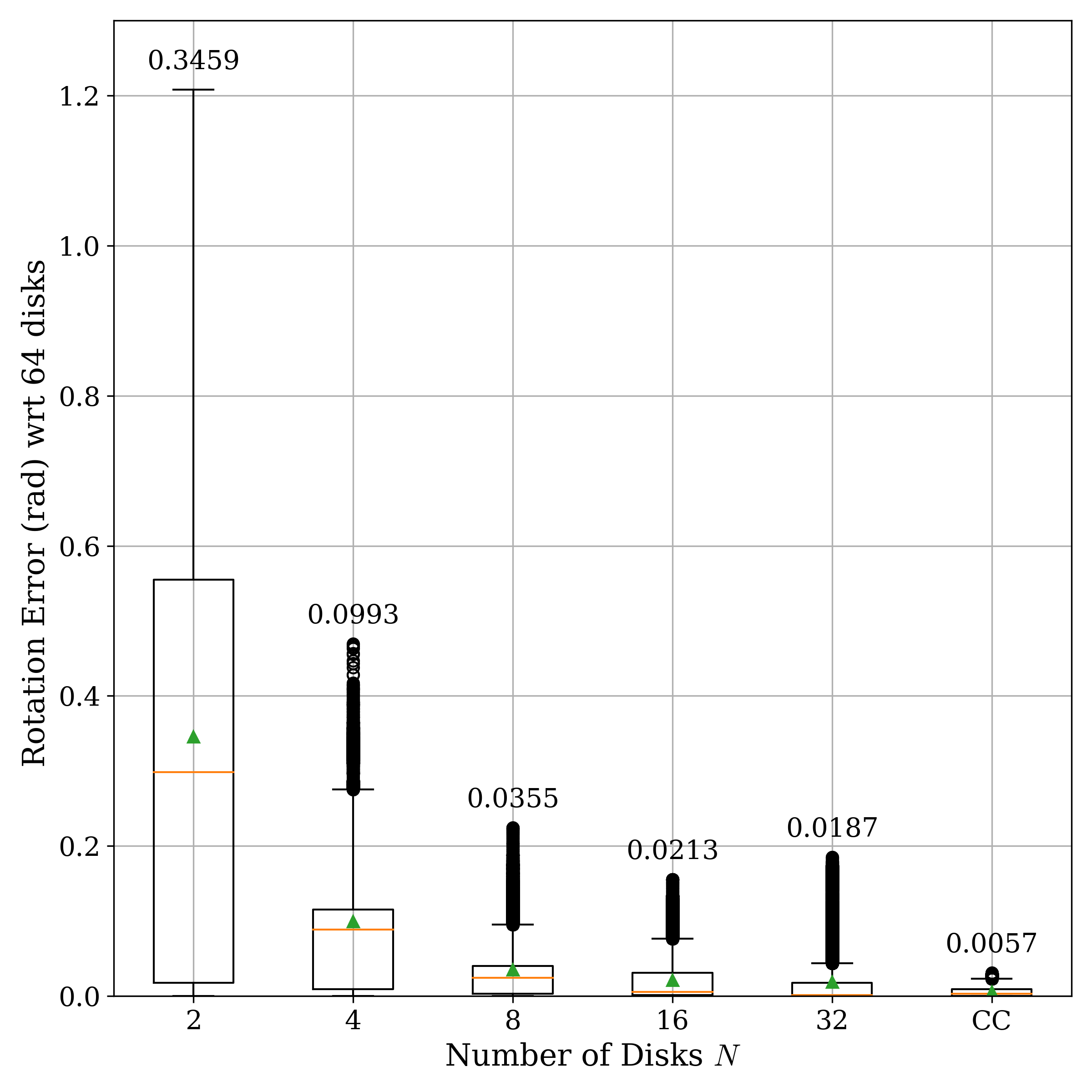}%
        \label{fig:tip_angle_error_64disks}%
    }
    \caption{Results of \textit{{Non-\ac{CC} vs \ac{UJ} Approximation}}: Tip position (a) and orientation (b) errors compared to a $N=64$ disk model as ground truth over the 60 second manipulation episode. The green triangles are means over the entire episode, and whose values are displayed above each box plot. CC indicates the constant curvature model.}
    \label{fig:64 disks error}
\end{figure*}

\textbf{Non-\ac{CC} vs \ac{UJ} Approximation:} Figure~\ref{fig:64 disks error} shows the results treating an $N=64$ disk simulation as ground truth. Each box plot is a set of data gathered over a single episode with a 5 kg box.
Recall that we use the \ac{CC} model to calculate configuration variables $q$ both in simulation and on hardware. The \ac{CC} assumption approximates true joint configuration with an average tip position error of 3 mm and orientation error of .0055 radians. This performance is satisfactory for large-scale grasping and lifting tasks. Hence, $q$ in the observation vector (Table~\ref{tab:observation_vector}) is suitable for policy learning. 

Interestingly, while the \ac{CC} orientation error outperforms all others, it is not the best in tip position, which has an average similar to $N=8$ disks. This is due to how the orientation is estimated. If the bottom and top plates are parallel, the \ac{CC} assumption dictates that the joint be perfectly straight, like a cylinder. Alternatively, if the bottom and top plates are parallel but the joint is not straight, as shown in Figure~\ref{fig:nonCC_approx}, the orientation errors will still be small, but the position errors will be higher. Specifically, point C in Figure~\ref{fig:nonCC_approx} would shift to the right. Hence, the tip \textit{position} error is the key indicator of how much non-\ac{CC} bending occurs. In this case, Figure~\ref{fig:tip_position_error_64disks} shows that the \ac{UJ} approximation with sufficiently large $N$ can approximate a true continuum joint better than the \ac{CC} model. 

Overall, Figures \ref{fig:realtime-factor}, \ref{fig:tip_errors}, and \ref{fig:64 disks error} demonstrate the typical tradeoff between speed and accuracy. For faster simulation, large time steps and few disks are desirable. But errors relative to both constant and non-\ac{CC} cases increase exponentially as $N$ decreases. If $N$ is too small, the sim-to-real gap becomes large enough to prevent successful deployment on hardware.

\subsection{Policy Training}
\begin{figure}[t]
    \centering
    \includegraphics[width=\columnwidth]{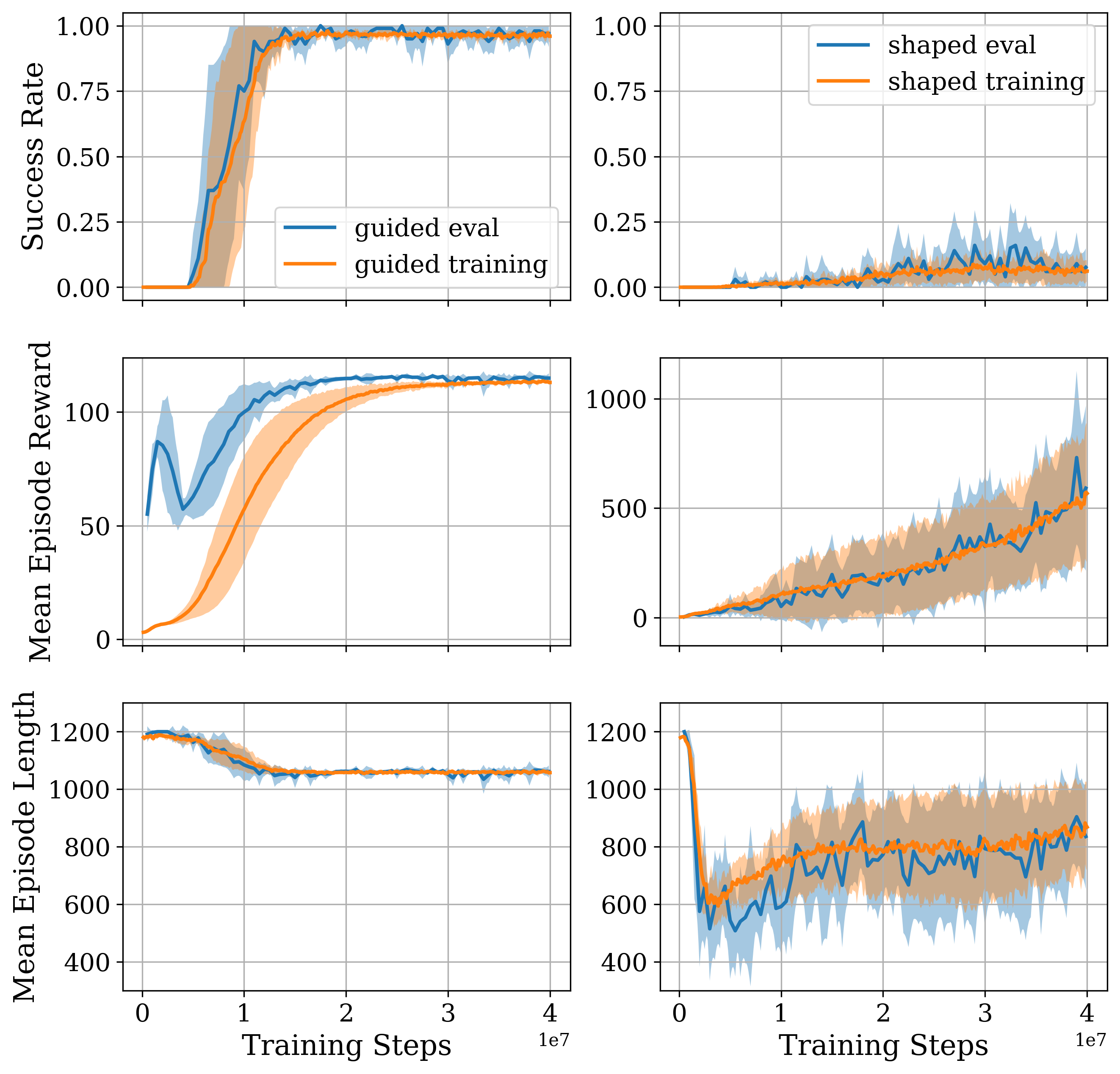}
    \caption{Average training metrics for training and evaluation averaged over 10 runs with different random seeds. The left column uses the guided reward term while the right column uses a nominal shaped reward based on previous work. Shaded areas indicate 1 standard deviation over all 10 runs. The success rate is clipped from 0 to 1.}
    \label{fig:training_metrics}
\end{figure} 

Figure~\ref{fig:training_metrics} presents the training metrics averaged across 10 random seeds for the guided reward (Eq.~\ref{eq:full-guided-reward}) in the left column and the shaped reward (Eq.~\ref{eq:full-shaped-reward}) in the right column. The training metrics include random exploration noise, while evaluation metrics are evaluated deterministically using the mean of the policy output. For the guided reward, the agent rapidly improves in the first 15 million training steps, where the success rate plateaus near 1.0, indicating that the task has largely been solved. The early discrepancies in episode reward between the rollout and evaluation statistics are due to the exploration noise during training. The drop in mean episode length indicates that the learned policy becomes increasingly efficient at completing the task.

Our comparison with a manually shaped reward function revealed significant challenges, including suboptimal final performance and learning instability. Despite a continually increasing reward signal throughout training, task success rate peaked early at 3E7 steps and failed to improve further. The quick drop in mean episode length was due to the agent learning undesirable behaviors, such as attempting premature grasps (i.e., grasping before the robot is close enough) that frequently led to tipping and episode termination. A notable characteristic of the shaped reward was the significantly higher variance in all signals, which points to a less stable learning process compared to that of the guided policy.

This finding illustrates a well-documented challenge in \ac{RL}: the sensitivity of policy performance to the precise formulation of the reward signal. The iterative process of designing and tuning reward functions can become computationally expensive and does not guarantee stable convergence to a desirable policy. Our guided method presents a compelling alternative. Providing a more structured exploration process via the motion primitive leads to improved final performance, enhanced sample efficiency, and greater training stability without the need for extensive reward engineering.

\begin{figure}[t]
    \centering
    \includegraphics[width=\columnwidth]{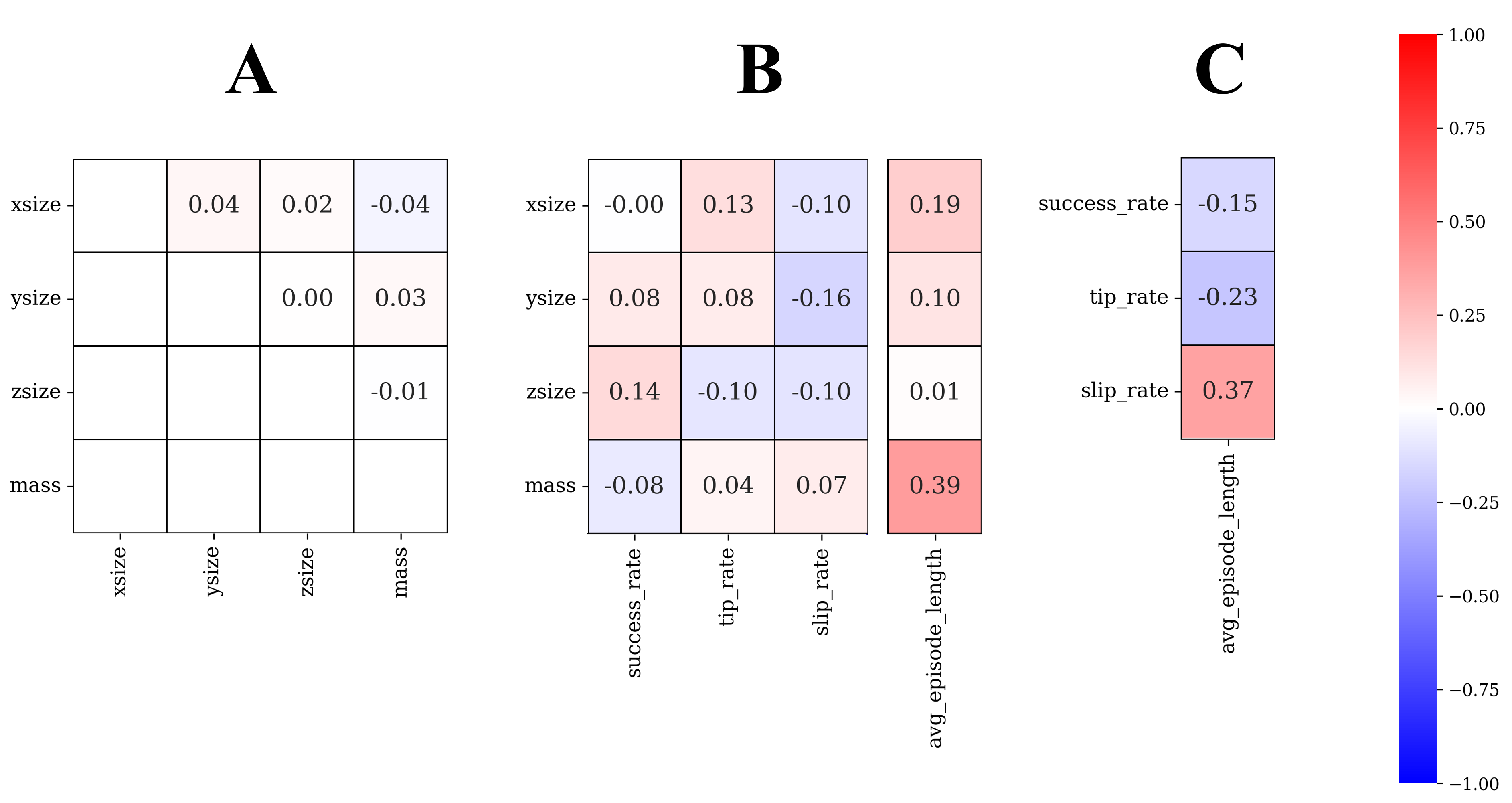}
    \caption{Spearman correlation coefficients between varied inputs and episode outcomes for 1000 test boxes in simulation. Each box was placed randomly in the robot's workspace the same as during training because we have observed that the initial pose of the box does affect episode outcome.}
    \label{fig:spearman-corr}
\end{figure}

\subsection{Policy Evaluation in Simulation}

\textbf{Object-wise Outcome Analysis:} We evaluated the trained policy on a new set of 1000 boxes with random properties sampled from the ranges in Table~\ref{tab:sim_params}. 
To achieve an unbiased policy evaluation, we increased the sample size up to 1000, until the Spearman correlation coefficients between the \texttt{xsize}, \texttt{ysize}, \texttt{zsize}, and \texttt{mass} were less than 0.05 (Figure~\ref{fig:spearman-corr}). The low correlation between object properties ensures a well-spread distribution of objects. The learned policy achieved a success rate of 94.1\%, a tip rate of 2.5\%, and a slip rate of 3.4\%. 
To further analyze how well the policy performs the task across different objects, we computed the correlation between object parameters (e.g., size, mass) and episode parameters (e.g., outcome, duration). Figure~\ref{fig:spearman-corr} shows the correlation matrix. We found that object properties are not strongly correlated with success rate. This indicates that the policy generalized well to the parameters in Table~\ref{tab:sim_params}. Among the object property correlations, the z-size is the strongest with $\rho = .14$, indicating that taller boxes are easier to lift. This is likely due to the safety limit we impose on the torso to avoid crushing the arms on the floor, so \blackout{Baloo} may not be able to grasp short boxes as well.
Moreover, a slight anti-correlation between the object dimensions and slip rate indicates that the policy is more likely to fail on smaller objects. This could be attributed to the large size of the robot, as we observed this pattern also in hardware.  
Conversely, the slip rate is positively correlated with average episode length ($\rho = .37)$. This is  because the episode is not terminated if the object slips, and continues up to 1200 steps unless the object is successfully lifted or tipped. 
Finally, heavier objects are more difficult to lift and generally take longer, as expected. Indeed, object mass is slightly anti-correlated with success rate ($\rho=-.08$), slightly correlated with tip/slip rates, and moderately correlated with average episode length ($\rho = .39$). 

\textbf{Motion Primitive vs Policy:} The results of this experiment are shown in Table~\ref{tab:baseline-transition-matrix}. The off-diagonal entries are cases where the outcome for a particular object is different for the motion primitive and the learned policy. The bottom row and right column are marginal totals for the learned policy and the motion primitive respectively. Using the Stuart-Maxwell test on this table, learning had a statistically significant effect with a p-value of 1.4e-2. 
Notably, policy learning increased the overall success compared to the motion primitive (from 931 to 941), as shown in the marginal totals. This is because the learned policy converted 13 slips and 6 tips of the motion primitive to success (see first column). However, the learned policy turned 9 successes with the motion primitive in 6 slips and 3 tips (see first row). Hence, learning did not have a universally positive impact. 
There is a notable pattern of slips being converted into tips. 10 objects that slipped with the motion primitive instead tipped with the learned policy. The reverse, from tip to slip, was not common (only 1 case). This suggests that in avoiding slips, the learned policy may sometimes be overly forceful, leading to a tip.
Overall, the learned policy demonstrates a clear and statistically significant improvement over the motion primitive. The primary benefits of learning were a higher overall success rate and a marked reduction in slip-related failures. However, the analysis also highlights areas for further improvement. The learned policy introduced a few failures on previously successful objects and showed a tendency to convert slip failures into tip failures.
%

\begin{table}[t]
\centering
\caption{Transition matrix comparing outcomes of the Primitive Policy (rows) vs. the Learned Policy (columns) on 1000 unperturbed test objects.}
\label{tab:baseline-transition-matrix}
\begin{tabular}{@{}l cccc@{}}
\toprule
& \multicolumn{4}{c}{\textbf{Learned Policy Outcome}} \\
\cmidrule(l){2-5} 
\textbf{Motion Primitive Outcome} & \textbf{Success} & \textbf{Slip} & \textbf{Tip} & \textbf{Total} \\ \midrule
\textbf{Success} & 922 & 6 & 3 & \textbf{931} \\
\textbf{Slip} & 13 & 27 & 10 & \textbf{50} \\
\textbf{Tip} & 6 & 1 & 12 & \textbf{19} \\ \midrule
\textbf{Total} & \textbf{941} & \textbf{34} & \textbf{25} & 1000 \\ \bottomrule
\end{tabular}
\end{table}

\textbf{Motion Primitive vs Policy (with perturbation):} The results of this experiment are shown in Table~\ref{tab:baseline-transition-matrix-perturbed}. Under the added stress of perturbations, the benefit of the learned, closed-loop policy became far more pronounced. The change in outcome distribution is highly statistically significant (p = 1.4e-12), demonstrating that the policy's reactive nature is a key advantage. 
The learned policy achieved 850 successes compared to the motion primitive's 822, a net increase of 28. The improvement is primarily driven by a 55\% reduction in slips. The motion primitive slipped on 112 objects, whereas the learned policy slipped on only 50. This confirms the policy can actively and effectively counteract the perturbation.
However, the net increase in success comes at a cost. Slips were reduced, but tips increased by 51\%, from 66 to 100. The Slip $\rightarrow$ Tip transition happened 48 times and is the largest off-diagonal cell in Table~\ref{tab:baseline-transition-matrix-perturbed}. This indicates that the policy's primary strategy to prevent slipping under perturbation is a correction that often results in the object tipping over.  

\begin{table}[t]
\centering
\caption{Transition matrix comparing outcomes of the Primitive Policy (rows) vs. the Learned Policy (columns) on 1000 test objects under a 100\,N downward perturbation.}
\label{tab:baseline-transition-matrix-perturbed}
\begin{tabular}{@{}l cccc@{}}
\toprule
& \multicolumn{4}{c}{\textbf{Learned Policy Outcome}} \\
\cmidrule(l){2-5} 
\textbf{Motion Primitive Outcome} & \textbf{Success} & \textbf{Slip} & \textbf{Tip} & \textbf{Total} \\ \midrule
\textbf{Success} & 815 & 2 & 5 & \textbf{822} \\
\textbf{Slip} & 20 & 44 & 48 & \textbf{112} \\
\textbf{Tip} & 15 & 4 & 47 & \textbf{66} \\ \midrule
\textbf{Total} & \textbf{850} & \textbf{50} & \textbf{100} & 1000 \\ \bottomrule
\end{tabular}
\end{table}

\begin{figure}[t]
    \centering
    \includegraphics[width=\columnwidth]{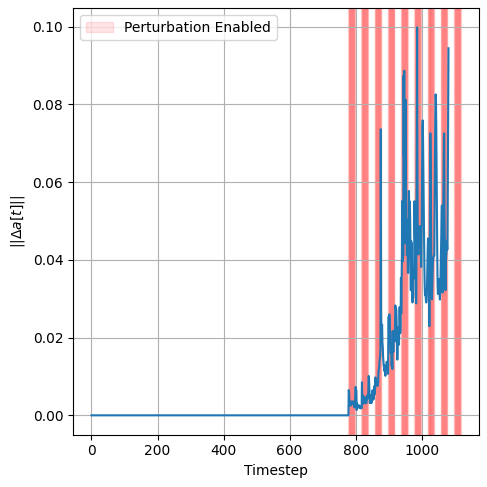}
    \caption{An example of a single trial that was Slip $\rightarrow$ Success, showing the perturbation pattern in red and the resulting corrective action in blue.}
    \label{fig:LOWRES}
\end{figure}

\textbf{Analysis of Policy Response to Perturbation:} To better understand the mechanism behind the differences in outcomes between Table~\ref{tab:baseline-transition-matrix} and \ref{tab:baseline-transition-matrix-perturbed}, we examined the policy's corrective actions. The analysis shows that the learned policy's actions change in direct response to perturbation and reveals how the nature of the change tends to affect the final outcome. The notation X $\rightarrow$ Y used in Table~\ref{tab:final-action-metrics} denotes a transition from outcome X to outcome Y.
Figure~\ref{fig:LOWRES} shows the corrective action as a percentage of the maximum action, for a sample of Slip $\rightarrow$ Success episode. Notice that, before the perturbation happens (as indicated by the vertical red bars), there is no corrective action. But immediately following the onset of perturbations, the actions change. While the actual corrective action trajectory varies between objects, the policy actively modulates its response with non-zero corrective actions after perturbation. This provides evidence that the closed-loop learned policy is reactive to external disturbances in a way that the open-loop motion primitive cannot.

\begin{table}[t]
\centering
\setlength{\tabcolsep}{4pt} 
\caption{Statistical analysis of the normalized corrective action magnitude ($||\Delta a||$) across different outcome transitions. The data reveals two distinct failure modes: (1) immediate, physically unsalvageable failures characterized by low peak force, and (2) unstable overcorrections characterized by a disproportionately large peak force.}
\label{tab:final-action-metrics}
\begin{tabular}{@{}l c c c c@{}}
\toprule
\textbf{Outcome Transition} & \textbf{\# of Trials} & \textbf{\begin{tabular}[c]{@{}c@{}}Mean\\ (\%)\end{tabular}} & \textbf{\begin{tabular}[c]{@{}c@{}}Median\\ (\%)\end{tabular}} & \textbf{\begin{tabular}[c]{@{}c@{}}Max\\ (\%)\end{tabular}} \\ \midrule
\multicolumn{5}{l}{\textit{Successful Grasps}} \\
\hspace{1em} Success $\rightarrow$ Success & 815 & 0.29 & 0.16 & 12.2 \\
\hspace{1em} Slip $\rightarrow$ Success & 20 & 0.40 & 0.21 & 4.8 \\
\hspace{1em} Tip $\rightarrow$ Success & 15 & 0.56 & 0.35 & 11.6 \\ \midrule
\multicolumn{5}{l}{\textit{Failed Grasps}} \\
\hspace{1em} Success $\rightarrow$ Tip & 5 & 0.85 & 0.62 & 3.6 \\
\hspace{1em} Success $\rightarrow$ Slip & 2 & 0.85 & 0.58 & 3.9 \\
\hspace{1em} Slip $\rightarrow$ Tip & 48 & 1.02 & 0.68 & 33.1 \\ \bottomrule
\end{tabular}
\end{table}

Table~\ref{tab:final-action-metrics} provides further insights into how the policy reactions affect the outcome. Successful grasps are characterized by small, controlled reactions. The Success $\rightarrow$ Success group required a mean corrective action of only 0.29\%. Even when saving a grasp that would have otherwise failed (Slip~$\rightarrow$~Success, Tip $\rightarrow$ Success), the policy's reaction remains measured, with mean corrective actions of 0.40\% and 0.56\%, respectively.
In contrast, the failed grasps are associated with more extreme responses that fall into two distinct categories: 

\begin{enumerate}
    \item \textit{Overcorrection:} The most common failure transition (Slip $\rightarrow$ Tip with 48 instances) is characterized by a massive corrective action. It has the highest mean effort (1.02\%) and the highest peak effort (33.1\%). When the policy detects a slip, it often over corrects and causes the box to tip over instead. 
    \item \textit{Insufficient Correction:} The Success $\rightarrow$ Tip and Success $\rightarrow$ Slip failure cases both have elevated averages of 0.85\% but their maximums are both surprisingly low at 3.6\% and 3.9\% respectively. This suggests that the box's state after the perturbation became unrecoverable so quickly that the policy failed to respond in time.
\end{enumerate}

In summary, this analysis reveals that the learned policy is reactive to external forces. It successfully applies small, modulated corrections to maintain stability. This can default to an overly aggressive response when faced with a significant slip, which can be addressed in future work.

\subsection{Zero-shot Sim-to-Real Transfer}
\label{sec:sim-to-real}

\begin{figure*}[]
    \centering
    \includegraphics[width=\textwidth]{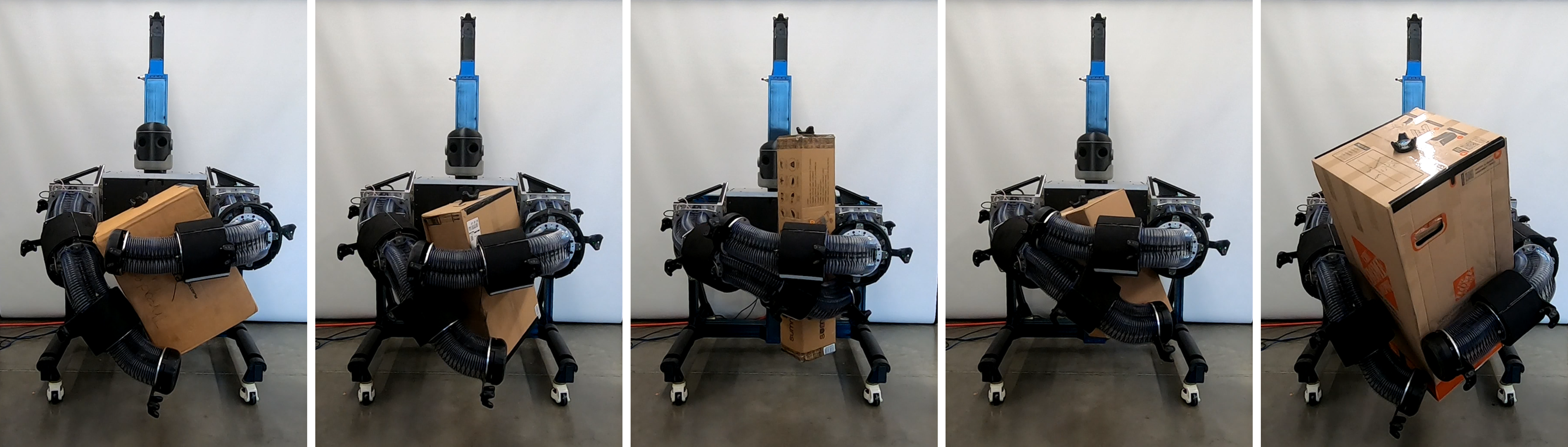}
    \caption{Hardware experiments for \texttt{Box\_1} (left) to \texttt{Box\_5} (right). Table~\ref{tab:experiment_results} summarizes the results from each of the 5 trials. Each box was attempted 5 times from different initial positions and orientations. The accompanying video\textsuperscript{\ref{videoFootnote}} shows all 25 attempts.}
    \label{fig:lift-all-boxes}
\end{figure*}

\begin{table}[t]
\centering
\caption{Sim-to-real whole-arm grasping results, reporting the number of trials for each outcome: tip, slip, and success.}
\label{tab:experiment_results}
\begin{tabular}{@{}llllll@{}}
\toprule
\textbf{Item}       & \textbf{Size (m)} & \textbf{Mass (kg)} & \textbf{Tip}      & \textbf{Slip} & \textbf{Success} \\ \midrule
\texttt{Box\_1} & 0.45 x 0.45 x 0.61 & 3.4 & 0 & 0 & 5  \\
\texttt{Box\_2} & 0.20 x 0.57 x 0.70 & 3.5 & 0 & 0 & 5  \\
\texttt{Box\_3} & 0.20 x 0.20 x 1.10 & 1.9 & 0 & 3 & 2  \\
\texttt{Box\_4} & 0.27 x 0.36 x 0.64 & 1.0 & 0 & 0 & 5  \\
\texttt{Box\_5} & 0.52 x 0.52 x 1.00 & 9.8 & 0 & 0 & 5  \\
\midrule
\multicolumn{3}{l}{\textbf{Aggregate results:}}  & 0/25 & 3/25 & 22/25\\
\bottomrule
\end{tabular}
\end{table}

We transferred the learned policy to the real \blackout{Baloo} without finetuning (zero-shot). As shown in Figure~\ref{fig:lift-all-boxes} and the accompanying video\footnote{\url{\videoUrl} \label{videoFootnote}}, the robot successfully lifted all 5 objects.
Table~\ref{tab:experiment_results} summarizes the sim-to-real transfer results, which align well with those from the simulation. Indeed, the success rate on hardware (88\%) closely matches the simulation result (94.1\%), with a 6.1\% gap. Slip is the second most frequent outcome in both settings, indicating good predictive alignment, although the slip rate is higher on hardware (12\% vs 3.4\%). Finally, the tip rate was low in both cases (0\% hardware, 2.5\% simulation), which is reasonable given the smaller hardware sample size of 25 vs 1000. 

The close alignment between simulation and hardware results supports the validation of our simulation approach as a useful tool in generating contact-rich manipulation strategies with soft robots. Remarkably, the policy transferred well to the real world without explicit sim-to-real transfer techniques or validation against experimental data before learning. We attribute this to our qualitative manual calibration of model parameters (i.e., masses, joint stiffness $K$, joint damping $C$) to obtain realistic simulations. 

In addition, the inherent compliance of \blackout{Baloo} facilitated the whole-body grasping task. This suggests that soft robotic structures possess inherent robustness to parametric uncertainties during grasping. This significant advantage over rigid counterparts reduces the need for extensive sim-to-real techniques such as domain randomization. Adopting soft robotic components could potentially accelerate deployment in unstructured, real-world environments where parametric uncertainty is unavoidable.

The policy is most likely to fail with smaller objects. For example, \texttt{Box\_3}, which slipped 3 times, had the minimum x and y dimensions of 20 cm, making it the most challenging object. We observed that the slips came from a self-collision between a link in the left arm and the end-effector cover on the right arm. The resultant jamming prevents the arms from closing fully on the object. We installed this protective cover on the hardware after training to avoid damage to the cardboard boxes. Since the simulation does not model the cover, it could explain the difference in slip rates between the simulation and hardware. As shown in the video\textsuperscript{\ref{videoFootnote}}, after a manual intervention, the arms close securely around \texttt{Box\_3}. This pattern was repeatable when \texttt{Box\_3} was positioned close to the chest. Conversely, when \texttt{Box\_3} was placed farther away, the arms typically grasp the object before jamming occurs. The same situation happens with the remaining boxes, which are larger than \texttt{Box\_3} in at least xsize or ysize.

\begin{figure*}
    \centering
    \includegraphics[width=\textwidth]{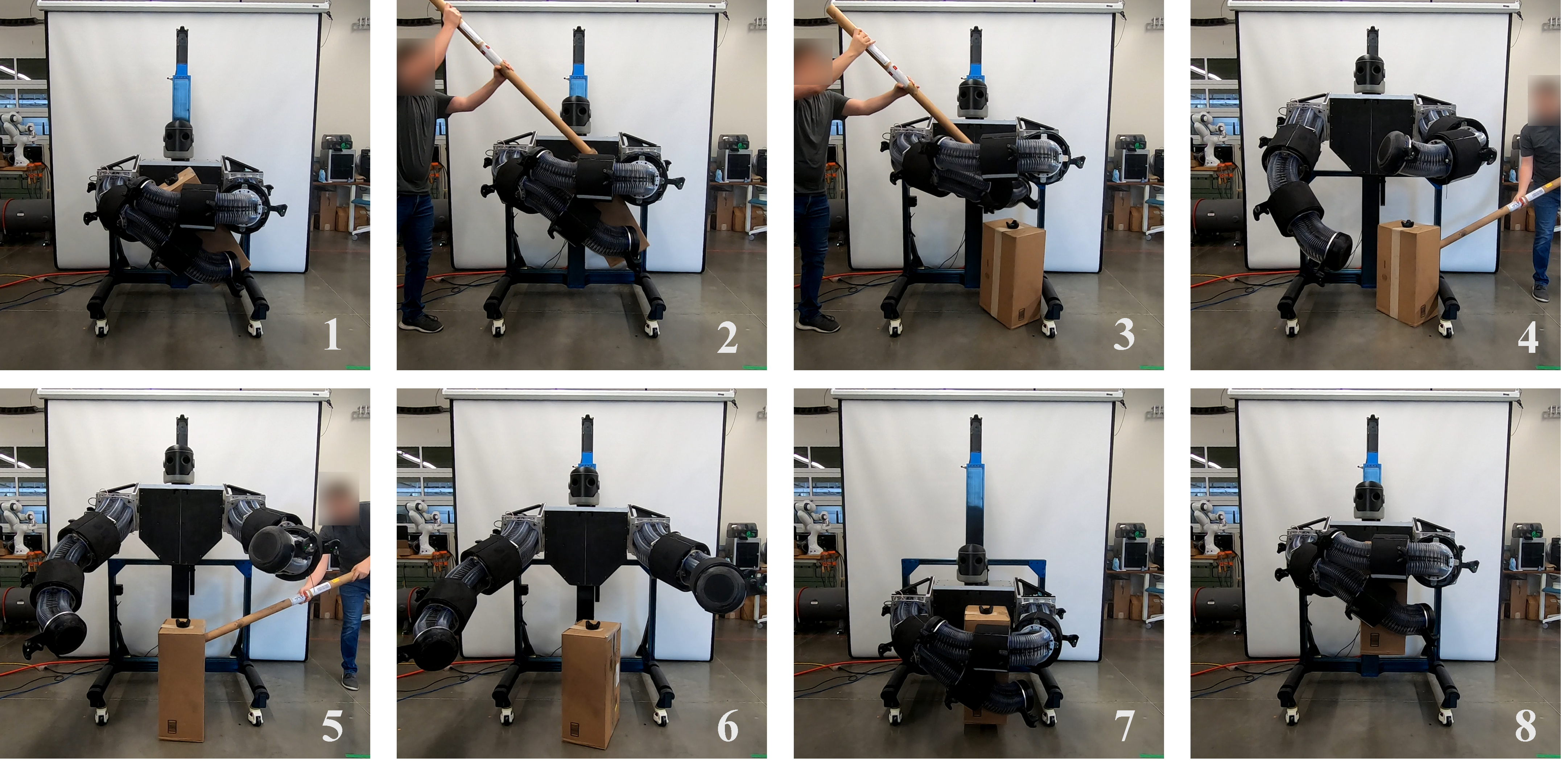}
    \caption{Emergent re-grasping behavior. After a successful lift in frame 1, we knock the box out of \blackout{Baloo's} grasp in frames 2-3 and push it farther away in frames 4-5. Frames 6-8 show the robot re-approaching and re-grasping the object.}
    \label{fig:regrasp}
\end{figure*}

\subsection{Emergent Whole-body Re-grasping Behavior}
A further result of the whole-body grasping policy is the re-grasping behavior. Notably, this behavior emerged without explicit encoding in the reward or the motion primitive (Algorithm~\ref{alg:open-loop-hugger}). Indeed, the original motion primitive used to guide training behaves as an inflexible state machine and will \textit{not} attempt to re-grasp the object. Thanks to the combination of policy learning and a reward informed by an open-loop motion trajectory, we achieved a reactive closed-loop controller capable of whole-body re-grasping. 

Figure~\ref{fig:regrasp} demonstrates the emergent re-grasping behavior when we forcefully hit \texttt{Box\_1} out of its grasp. First, we let the policy autonomously lift the box. Then, we hit the box out of grasp and push it to a new location. As a result, the robot opened its arms and approached the object again, achieving another successful lift. The same behavior happened with \texttt{Box\_3}, as shown in the video\footnotemark[\getrefnumber{videoFootnote}]. In this case, rather than manually pushing the object out of \blackout{Baloo}'s grasp, it slipped due to self-collision (discussed previously). Despite the slip, the robot could still lift the object when a small irregularity in the box caught onto the arm. As the robot lifted the box, however, the measured box position did not match what was expected for a box of that size, which is in the observation space (Table~\ref{tab:observation_vector}). Therefore, the arms dropped the object and tried again. We could attribute this to the ability of the policy to infer grasp slippage. Since the self-collision is so repeatable, this reattempt cycle continues until the box falls into a location where self-collision does not happen.

\section{Conclusion}
\label{sec:conclusion}

In this work, we have made significant strides in addressing the outstanding challenge of whole-body manipulation of large, heavy objects with rigid-soft robots. Our approach centers on three key contributions. First, we developed and have open-sourced a novel, high-speed simulation framework for contact-rich manipulation with soft robots, built upon MuJoCo. This framework is not only fast enough for data-intensive reinforcement learning, running up to 350x real-time on a single thread, but also accurate enough to enable successful zero-shot transfer of learned policies. Second, utilizing this simulation, we achieved an impressive 88\% success rate in zero-shot sim-to-real transfer for a complex, whole-body grasping policy, a first for coordinated, large-scale soft robot arms. This success was enabled by our open-source RL pipeline, which uses a guided reinforcement learning approach with a motion primitive, proving critical where standard methods fail. Finally, through an in-depth analysis, we demonstrated that this guided learning process transforms the initial open-loop primitive into a fundamentally new, reactive, closed-loop policy. This learned reactivity enables advanced behaviors such as re-grasping, showcasing a significant advancement in creating intelligent and adaptive soft robotic systems.

\section{Limitations and Future Work}
A limitation of our guided reinforcement learning method is its reliance on a hand-designed motion primitive to begin the learning process. Although this is less demanding than providing full teleoperated demonstrations, it does mean that a new primitive must be designed for each new task. Future work could investigate methods for learning these primitives from a small number of non-expert demonstrations or for chaining learned primitives to handle more complex, multi-stage tasks, which would enhance the versatility of our approach. A systematic study on how the quality and variations of the motion primitive influence the learned policy would also be a valuable contribution.
Furthermore, our experiments revealed that the learned reactive policy can lead to complex and dynamic failure modes, such as forceful over-correction when faced with unknown disturbances. This highlights the challenge of teaching a policy not just to react, but to react appropriately. The policy's reactive capabilities could be enhanced by incorporating perturbations into the training process, enabling it to learn more robust and nuanced recovery strategies. However, this introduces the challenge of designing an effective curriculum that balances task difficulty with learning stability.

Many of the observed hardware failures were also due to the robot jamming against itself during the initial approach, highlighting the need for a more refined pre-grasp/approach strategy. Similarly, the absence of force/tactile feedback led to some objects being grasped too tightly. Future work will focus on integrating tactile sensing into the learning pipeline to improve grasp stability and encourage more delicate handling of objects, while also explicitly learning an optimal pre-grasp approach to mitigate jamming issues. To further address the discrepancies between simulation and reality, a systematic investigation to fully close this sim-to-real gap could involve higher-fidelity modeling and conducting a sensitivity analysis on uncertain dynamics, such as friction and the non-uniform mass distribution of the weighted cardboard boxes.

Finally, our current system relies on an external motion tracking system for object pose and robot proprioception. We plan to explore vision-based estimation of the object's pose and the possibility of learning directly from visual data in future iterations.


%



\section*{Acknowledgment}

This work was supported by the \blackout{National Science Foundation under Grant No. 1935312}.

\ifCLASSOPTIONcaptionsoff
  \newpage
\fi



\bibliographystyle{IEEEtran}

\bibliography{IEEEabrv, main}

\begin{thebibliography}{10}
\providecommand{\url}[1]{#1}
\csname url@samestyle\endcsname
\providecommand{\newblock}{\relax}
\providecommand{\bibinfo}[2]{#2}
\providecommand{\BIBentrySTDinterwordspacing}{\spaceskip=0pt\relax}
\providecommand{\BIBentryALTinterwordstretchfactor}{4}
\providecommand{\BIBentryALTinterwordspacing}{\spaceskip=\fontdimen2\font plus
\BIBentryALTinterwordstretchfactor\fontdimen3\font minus \fontdimen4\font\relax}
\providecommand{\BIBforeignlanguage}[2]{{%
\expandafter\ifx\csname l@#1\endcsname\relax
\typeout{** WARNING: IEEEtran.bst: No hyphenation pattern has been}%
\typeout{** loaded for the language `#1'. Using the pattern for}%
\typeout{** the default language instead.}%
\else
\language=\csname l@#1\endcsname
\fi
#2}}
\providecommand{\BIBdecl}{\relax}
\BIBdecl

\bibitem{rus2015design}
D.~Rus and M.~T. Tolley, ``Design, fabrication and control of soft robots,'' \emph{Nature}, vol. 521, no. 7553, pp. 467--475, 2015.

\bibitem{wang2022control}
J.~Wang and A.~Chortos, ``Control strategies for soft robot systems,'' \emph{Advanced Intelligent Systems}, vol.~4, no.~5, p. 2100165, 2022.

\bibitem{della2023model}
C.~Della~Santina, C.~Duriez, and D.~Rus, ``Model-based control of soft robots: A survey of the state of the art and open challenges,'' \emph{IEEE Control Systems Magazine}, vol.~43, no.~3, pp. 30--65, 2023.

\bibitem{falotico2024learning}
E.~Falotico, E.~Donato, C.~Alessi, E.~Setti, M.~S. Nazeer, C.~Agabiti, D.~Caradonna, D.~Bianchi, F.~Piqué, Y.~T. Ansari, and M.~Killpack, ``Learning controllers for continuum soft manipulators: Impact of modeling and looming challenges,'' \emph{Advanced Intelligent Systems}, vol. n/a, no. n/a, p. 2400344, 2024.

\bibitem{ibarz2021train}
J.~Ibarz, J.~Tan, C.~Finn, M.~Kalakrishnan, P.~Pastor, and S.~Levine, ``How to train your robot with deep reinforcement learning: lessons we have learned,'' \emph{The International Journal of Robotics Research}, vol.~40, no. 4-5, pp. 698--721, 2021.

\bibitem{todorov2012mujoco}
E.~Todorov, T.~Erez, and Y.~Tassa, ``Mujoco: A physics engine for model-based control,'' in \emph{2012 IEEE/RSJ international conference on intelligent robots and systems}.\hskip 1em plus 0.5em minus 0.4em\relax IEEE, 2012, pp. 5026--5033.

\bibitem{Barreiros_Onol_Zhang_Creasey_Goncalves_Beaulieu_Bhat_Tsui_Alspach_2025}
J.~A. Barreiros, A.~O. Onol, M.~Zhang, S.~Creasey, A.~Goncalves, A.~Beaulieu, A.~Bhat, K.~M. Tsui, and A.~Alspach, ``\BIBforeignlanguage{en}{Learning contact-rich whole-body manipulation with example-guided reinforcement learning},'' \emph{\BIBforeignlanguage{en}{Science RoBoticS}}, 2025.

\bibitem{armanini2023soft}
C.~Armanini, F.~Boyer, A.~T. Mathew, C.~Duriez, and F.~Renda, ``Soft robots modeling: A structured overview,'' \emph{IEEE Transactions on Robotics}, 2023.

\bibitem{webster2010design}
R.~J. Webster~III and B.~A. Jones, ``Design and kinematic modeling of constant curvature continuum robots: A review,'' \emph{The International Journal of Robotics Research}, vol.~29, no.~13, pp. 1661--1683, 2010.

\bibitem{faure2012sofa}
F.~Faure, C.~Duriez, H.~Delingette, J.~Allard, B.~Gilles, S.~Marchesseau, H.~Talbot, H.~Courtecuisse, G.~Bousquet, I.~Peterlik \emph{et~al.}, ``Sofa: A multi-model framework for interactive physical simulation,'' \emph{Soft tissue biomechanical modeling for computer assisted surgery}, pp. 283--321, 2012.

\bibitem{Coevoet_Morales-Bieze_Largilliere_Zhang_Thieffry_Sanz-Lopez_Carrez_Marchal_Goury_Dequidt}
E.~Coevoet, T.~Morales-Bieze, F.~Largilliere, Z.~Zhang, M.~Thieffry, M.~Sanz-Lopez, B.~Carrez, D.~Marchal, O.~Goury, J.~Dequidt, and C.~Duriez, ``\BIBforeignlanguage{en}{Software toolkit for modeling, simulation, and control of soft robots},'' \emph{\BIBforeignlanguage{en}{Advanced Robotics}}, vol.~31, no.~22, p. 1208–1224, Nov. 2017.

\bibitem{dubied2022sim}
M.~Dubied, M.~Y. Michelis, A.~Spielberg, and R.~K. Katzschmann, ``Sim-to-real for soft robots using differentiable fem: Recipes for meshing, damping, and actuation,'' \emph{IEEE Robotics and Automation Letters}, vol.~7, no.~2, pp. 5015--5022, 2022.

\bibitem{alessi2023ablation}
C.~Alessi, E.~Falotico, and A.~Lucantonio, ``Ablation study of a dynamic model for a 3d-printed pneumatic soft robotic arm,'' \emph{IEEE Access}, 2023.

\bibitem{Katzschmann_Santina_Toshimitsu_Bicchi_Rus_2019}
\BIBentryALTinterwordspacing
R.~K. Katzschmann, C.~D. Santina, Y.~Toshimitsu, A.~Bicchi, and D.~Rus, ``\BIBforeignlanguage{en}{Dynamic motion control of multi-segment soft robots using piecewise constant curvature matched with an augmented rigid body model},'' in \emph{\BIBforeignlanguage{en}{2019 2nd IEEE International Conference on Soft Robotics (RoboSoft)}}.\hskip 1em plus 0.5em minus 0.4em\relax Seoul, Korea (South): IEEE, Apr. 2019, p. 454–461. [Online]. Available: \url{https://ieeexplore.ieee.org/document/8722799/}
\BIBentrySTDinterwordspacing

\bibitem{graule2021somo}
M.~A. Graule, C.~B. Teeple, T.~P. McCarthy, G.~R. Kim, R.~C.~S. Louis, and R.~J. Wood, ``Somo: Fast and accurate simulations of continuum robots in complex environments,'' in \emph{2021 IEEE/RSJ International Conference on Intelligent Robots and Systems (IROS)}.\hskip 1em plus 0.5em minus 0.4em\relax IEEE, 2021, pp. 3934--3941.

\bibitem{Schegg_Ménager_Khairallah_Marchal_Dequidt_Preux_Duriez_2023}
P.~Schegg, E.~Ménager, E.~Khairallah, D.~Marchal, J.~Dequidt, P.~Preux, and C.~Duriez, ``\BIBforeignlanguage{en}{Sofagym: An open platform for reinforcement learning based on soft robot simulations},'' \emph{\BIBforeignlanguage{en}{Soft Robotics}}, vol.~10, no.~2, p. 410–430, Apr. 2023.

\bibitem{Morimoto_Ikeda_Niiyama_Kuniyoshi_2022}
R.~Morimoto, M.~Ikeda, R.~Niiyama, and Y.~Kuniyoshi, ``\BIBforeignlanguage{en}{Characterization of continuum robot arms under reinforcement learning and derived improvements},'' \emph{\BIBforeignlanguage{en}{Frontiers in Robotics and AI}}, vol.~9, p. 895388, Sep. 2022.

\bibitem{Morimoto_Nishikawa_Niiyama_Kuniyoshi_2021}
\BIBentryALTinterwordspacing
R.~Morimoto, S.~Nishikawa, R.~Niiyama, and Y.~Kuniyoshi, ``Model-free reinforcement learning with ensemble for a soft continuum robot arm,'' in \emph{2021 IEEE 4th International Conference on Soft Robotics (RoboSoft)}, Apr. 2021, p. 141–148. [Online]. Available: \url{https://ieeexplore.ieee.org/document/9479340/?arnumber=9479340}
\BIBentrySTDinterwordspacing

\bibitem{naughton2021elastica}
N.~Naughton, J.~Sun, A.~Tekinalp, T.~Parthasarathy, G.~Chowdhary, and M.~Gazzola, ``Elastica: A compliant mechanics environment for soft robotic control,'' \emph{IEEE Robotics and Automation Letters}, vol.~6, no.~2, pp. 3389--3396, 2021.

\bibitem{Jitosho_Lum_Okamura_Liu_2023}
\BIBentryALTinterwordspacing
R.~Jitosho, T.~G.~W. Lum, A.~Okamura, and K.~Liu, ``\BIBforeignlanguage{en}{Reinforcement learning enables real-time planning and control of agile maneuvers for soft robot arms},'' in \emph{\BIBforeignlanguage{en}{Proceedings of The 7th Conference on Robot Learning}}.\hskip 1em plus 0.5em minus 0.4em\relax PMLR, Dec. 2023, p. 1131–1153. [Online]. Available: \url{https://proceedings.mlr.press/v229/jitosho23a.html}
\BIBentrySTDinterwordspacing

\bibitem{agabiti2023whole}
C.~Agabiti, E.~M{\'e}nager, and E.~Falotico, ``Whole-arm grasping strategy for soft arms to capture space debris,'' in \emph{2023 IEEE International Conference on Soft Robotics (RoboSoft)}.\hskip 1em plus 0.5em minus 0.4em\relax IEEE, 2023, pp. 1--6.

\bibitem{alessi2024pushing}
C.~Alessi, D.~Bianchi, G.~Stano, M.~Cianchetti, and E.~Falotico, ``Pushing with soft robotic arms via deep reinforcement learning,'' \emph{Advanced Intelligent Systems}, 2024.

\bibitem{zwane2024learning}
S.~Zwane, D.~Cheney, C.~C. Johnson, Y.~Luo, Y.~Bekiroglu, M.~D. Killpack, and M.~P. Deisenroth, ``Learning dynamic tasks on a large-scale soft robot in a handful of trials,'' in \emph{2024 IEEE/RSJ International Conference on Intelligent Robots and Systems (IROS)}.\hskip 1em plus 0.5em minus 0.4em\relax IEEE, 2024, pp. 11\,388--11\,393.

\bibitem{Graule_McCarthy_Teeple_Werfel_Wood_2022}
M.~A. Graule, T.~P. McCarthy, C.~B. Teeple, J.~Werfel, and R.~J. Wood, ``Somogym: A toolkit for developing and evaluating controllers and reinforcement learning algorithms for soft robots,'' \emph{IEEE Robotics and Automation Letters}, vol.~7, no.~2, p. 4071–4078, Apr. 2022.

\bibitem{zare2024survey}
M.~Zare, P.~M. Kebria, A.~Khosravi, and S.~Nahavandi, ``A survey of imitation learning: Algorithms, recent developments, and challenges,'' \emph{IEEE Transactions on Cybernetics}, 2024.

\bibitem{ravichandar2020recent}
H.~Ravichandar, A.~S. Polydoros, S.~Chernova, and A.~Billard, ``Recent advances in robot learning from demonstration,'' \emph{Annual review of control, robotics, and autonomous systems}, vol.~3, no.~1, pp. 297--330, 2020.

\bibitem{sarker2025review}
A.~Sarker, T.~Ul~Islam, and M.~R. Islam, ``A review on recent trends of bioinspired soft robotics: Actuators, control methods, materials selection, sensors, challenges, and future prospects,'' \emph{Advanced Intelligent Systems}, vol.~7, no.~3, p. 2400414, 2025.

\bibitem{nazeer2023soft}
M.~S. Nazeer, C.~Laschi, and E.~Falotico, ``Soft dagger: Sample-efficient imitation learning for control of soft robots,'' \emph{Sensors}, vol.~23, no.~19, p. 8278, 2023.

\bibitem{DellaSantina_Katzschmann_Biechi_Rus_2018}
\BIBentryALTinterwordspacing
C.~Della~Santina, R.~K. Katzschmann, A.~Biechi, and D.~Rus, ``\BIBforeignlanguage{en}{Dynamic control of soft robots interacting with the environment},'' in \emph{\BIBforeignlanguage{en}{2018 IEEE International Conference on Soft Robotics (RoboSoft)}}.\hskip 1em plus 0.5em minus 0.4em\relax Livorno: IEEE, Apr. 2018, p. 46–53. [Online]. Available: \url{https://ieeexplore.ieee.org/document/8404895/}
\BIBentrySTDinterwordspacing

\bibitem{Ma_Monk_Cheneler_2022}
\BIBentryALTinterwordspacing
N.~Ma, S.~Monk, and D.~Cheneler, ``\BIBforeignlanguage{en}{Design, prototyping and test of a dual-arm continuum robot for underwater environments},'' in \emph{\BIBforeignlanguage{en}{2022 7th International Conference on Robotics and Automation Engineering (ICRAE)}}.\hskip 1em plus 0.5em minus 0.4em\relax Singapore: IEEE, Nov. 2022, p. 158–164. [Online]. Available: \url{https://ieeexplore.ieee.org/document/10054612/}
\BIBentrySTDinterwordspacing

\bibitem{Wang_Xu_2019}
Y.~Wang and Q.~Xu, ``\BIBforeignlanguage{en}{Design and fabrication of a new dual-arm soft robotic manipulator},'' \emph{\BIBforeignlanguage{en}{Actuators}}, vol.~8, no.~1, p.~5, Jan. 2019.

\bibitem{Oh_Rodrigue}
N.~Oh and H.~Rodrigue, ``\BIBforeignlanguage{en}{Toward the development of large-scale inflatable robotic arms using hot air welding}.''

\bibitem{Russo_Sriratanasak_Ba_Dong_Mohammad_Axinte_2022}
M.~Russo, N.~Sriratanasak, W.~Ba, X.~Dong, A.~Mohammad, and D.~Axinte, ``\BIBforeignlanguage{en}{Cooperative continuum robots: Enhancing individual continuum arms by reconfiguring into a parallel manipulator},'' \emph{\BIBforeignlanguage{en}{IEEE Robotics and Automation Letters}}, vol.~7, no.~2, p. 1558–1565, Apr. 2022.

\bibitem{Kraus_Jensen_Killpack_2020}
\BIBentryALTinterwordspacing
D.~Kraus, A.~Jensen, and M.~Killpack, ``\BIBforeignlanguage{en}{Coordinated soft robot multi-arm manipulation},'' in \emph{\BIBforeignlanguage{en}{2020 3rd IEEE International Conference on Soft Robotics (RoboSoft)}}.\hskip 1em plus 0.5em minus 0.4em\relax New Haven, CT, USA: IEEE, May 2020, p. 424–431. [Online]. Available: \url{https://ieeexplore.ieee.org/document/9116029/}
\BIBentrySTDinterwordspacing

\bibitem{johnson2024baloolargescalehybridsoft}
\BIBentryALTinterwordspacing
C.~C. Johnson, A.~Clawson, and M.~D. Killpack, ``Baloo: A large-scale hybrid soft robotic torso for whole-arm manipulation,'' 2024. [Online]. Available: \url{https://arxiv.org/abs/2409.08420}
\BIBentrySTDinterwordspacing

\bibitem{pneudrive}
C.~C. Johnson, D.~G. Cheney, D.~L. Cordon, and M.~D. Killpack, ``Pneudrive: An embedded pressure control system and modeling toolkit for large-scale soft robots,'' in \emph{2024 IEEE 7th International Conference on Soft Robotics (RoboSoft)}, 2024, pp. 786--792.

\bibitem{Allen_Rupert_Duggan_Hein_Albert_2020}
T.~F. Allen, L.~Rupert, T.~R. Duggan, G.~Hein, and K.~Albert, ``Closed-form non-singular constant-curvature continuum manipulator kinematics,'' in \emph{2020 3rd IEEE International Conference on Soft Robotics (RoboSoft)}, New Haven, CT, USA, May 2020, p. 410–416.

\bibitem{hyatt2018configuration}
P.~Hyatt, D.~Kraus, V.~Sherrod, L.~Rupert, N.~Day, and M.~D. Killpack, ``Configuration estimation for accurate position control of large-scale soft robots,'' \emph{IEEE/ASME Transactions on Mechatronics}, vol.~24, no.~1, pp. 88--99, 2018.

\bibitem{berscheid2021jerk}
L.~Berscheid and T.~Kr{\"o}ger, ``Jerk-limited real-time trajectory generation with arbitrary target states,'' \emph{Robotics: Science and Systems XVII}, 2021.

\bibitem{pardo2022timelimitsreinforcementlearning}
\BIBentryALTinterwordspacing
F.~Pardo, A.~Tavakoli, V.~Levdik, and P.~Kormushev, ``Time limits in reinforcement learning,'' 2022. [Online]. Available: \url{https://arxiv.org/abs/1712.00378}
\BIBentrySTDinterwordspacing

\bibitem{faramaDeepDive}
``Deep {D}ive: {G}ymnasium {S}tep {A}{P}{I},'' \url{https://farama.org/Gymnasium-Terminated-Truncated-Step-API}, 2023, [Accessed 22-04-2025].

\bibitem{hussein2017imitation}
A.~Hussein, M.~M. Gaber, E.~Elyan, and C.~Jayne, ``Imitation learning: A survey of learning methods,'' \emph{ACM Computing Surveys (CSUR)}, vol.~50, no.~2, pp. 1--35, 2017.

\bibitem{Zhang_Barreiros_Onol_2023}
\BIBentryALTinterwordspacing
M.~Zhang, J.~Barreiros, and A.~O. Onol, ``Plan-guided reinforcement learning for whole-body manipulation,'' no. arXiv:2310.12263, Oct. 2023, arXiv:2310.12263 [cs]. [Online]. Available: \url{http://arxiv.org/abs/2310.12263}
\BIBentrySTDinterwordspacing

\bibitem{8463162}
A.~Nair, B.~McGrew, M.~Andrychowicz, W.~Zaremba, and P.~Abbeel, ``Overcoming exploration in reinforcement learning with demonstrations,'' in \emph{2018 IEEE International Conference on Robotics and Automation (ICRA)}, 2018, pp. 6292--6299.

\bibitem{pertsch2021skild}
K.~Pertsch, Y.~Lee, Y.~Wu, and J.~J. Lim, ``Demonstration-guided reinforcement learning with learned skills,'' \emph{5th Conference on Robot Learning}, 2021.

\bibitem{brys2015reinforcement}
T.~Brys, A.~Harutyunyan, H.~B. Suay, S.~Chernova, M.~E. Taylor, and A.~Now{\'e}, ``Reinforcement learning from demonstration through shaping.'' in \emph{IJCAI}, 2015, pp. 3352--3358.

\bibitem{stable-baselines3}
\BIBentryALTinterwordspacing
A.~Raffin, A.~Hill, A.~Gleave, A.~Kanervisto, M.~Ernestus, and N.~Dormann, ``Stable-baselines3: Reliable reinforcement learning implementations,'' \emph{Journal of Machine Learning Research}, vol.~22, no. 268, pp. 1--8, 2021. [Online]. Available: \url{http://jmlr.org/papers/v22/20-1364.html}
\BIBentrySTDinterwordspacing

\bibitem{Vulin_Christen_Stevšić_Hilliges_2021}
N.~Vulin, S.~Christen, S.~Stevšić, and O.~Hilliges, ``Improved learning of robot manipulation tasks via tactile intrinsic motivation,'' \emph{IEEE Robotics and Automation Letters}, vol.~6, no.~2, p. 2194–2201, Apr. 2021.

\bibitem{Raza}
M.~Raza, ``\BIBforeignlanguage{en}{Whole-body manipulation using reinforcement learning}.''

\end{thebibliography}

\end{document}